\documentclass[10pt,twocolumn,letterpaper]{article}

\usepackage[pagenumbers]{wacv} %

\def\confName{WACV}
\def\confYear{2026}

\def\paperTitle{Streaming Real-Time Trajectory Prediction Using Endpoint-Aware Modeling}

\def\authorBlock{
    Alexander Prutsch \qquad %
    David Schinagl \qquad $\;$ 
    Horst Possegger $\;$  \\
    Institute of Visual Computing, Graz University of Technology \\
    {\tt\small \{alexander.prutsch, david.schinagl, possegger\}@tugraz.at}
}

\newif\ifreview 
\newif\ifarxiv 
\newif\ifcamera 
\newif\ifrebuttal

\usepackage{graphicx}	
\usepackage{amsmath}	
\usepackage{amssymb}	
\usepackage{booktabs}
\usepackage{times}
\usepackage{microtype}
\usepackage{epsfig}
\usepackage{caption}
\usepackage{float}
\usepackage{placeins}
\usepackage{color, colortbl}
\usepackage{stfloats}
\usepackage{enumitem}
\usepackage{tabularx}
\usepackage{xstring}
\usepackage{multirow}
\usepackage{xspace}
\usepackage{url}
\usepackage{subcaption}
\usepackage{xcolor}
\usepackage[hang,flushmargin]{footmisc}
\usepackage{pifont}
\usepackage{tikz}
\usetikzlibrary[positioning, fit]

\ifcamera \usepackage[accsupp]{axessibility} \fi

\newcommand{\nbf}[1]{{\noindent \textbf{#1.}}}

\newcommand{\bval}[1]{\textbf{#1}}
\newcommand{\sval}[1]{\underline{#1}}
\newcommand{\seconds}[1]{$#1\,\mathrm{s}$}

\ifarxiv  \fi

\definecolor{rcol}{rgb}{0.9,0.9,0.9}

\newcommand{\R}[1]{{%
    \textbf{%
        \ifstrequal{#1}{1}{\textcolor{teal}{KqXf17}}{%
        \ifstrequal{#1}{2}{\textcolor{blue}{NnAZ16}}{%
        \ifstrequal{#1}{3}{\textcolor{olive}{2QBW15}}{%
        \ifstrequal{#1}{4}{\textcolor{teal}{R#1}}{%
                           \textcolor{cyan}{R#1}%
        }}}}%
    }%
}}

\newcommand{\mn}{SEAM}
\newcommand{\cmark}{\ding{51}}%
\newcommand{\xmark}{\color{gray}{\ding{55}}}

\newcount\mycount
\newread\myfile
    
\newcommand{\addresrow}[9]{

    \edef\temp{#9}

    \ifdefempty{\temp}{
        \node (#1_0) {\includegraphics[trim={#3cm, #4cm, #5cm, #6cm}, clip, width=7cm]{figures/raw_results_split/#2_0.pdf}};
    }{
        \node (#1_0) [below=of #9_0, yshift=-0.9cm] {\includegraphics[trim={#3cm, #4cm, #5cm, #6cm}, clip, width=7cm]{figures/raw_results_split/#2_0.pdf}};
    }

    \node (#1_1) [right=of #1_0, xshift=#7*7cm] {\includegraphics[trim={#3cm, #4cm, #5cm, #6cm}, clip, width=7cm]{figures/raw_results_split/#2_1.pdf}};
    \node (#1_2) [right=of #1_1, xshift=#7*7cm] {\includegraphics[trim={#3cm, #4cm, #5cm, #6cm}, clip, width=7cm]{figures/raw_results_split/#2_2.pdf}};
    
    \node (#1_3) [right=of #1_2, xshift=#7*7+0.6cm] {\includegraphics[trim={#3cm, #4cm, #5cm, #6cm}, clip, width=7cm]{figures/raw_results_split/#2_remo.pdf}};

    \ifdefempty{\temp}{
        \draw[thick, gray] ([xshift=0.3cm]#1_2.south east) -- ([xshift=0.3cm, yshift=0.5cm]#1_2.north east);
    }{
        \draw[thick, gray] ([xshift=0.3cm]#1_2.south east) -- ([xshift=0.3cm]#9_2.south east);
    }
    
    \mycount=0

    \openin\myfile=figures/raw_results_split/#2.txt
    \loop
        \unless\ifeof\myfile
        \read\myfile to \linecontent
        \ifnum\mycount<4
            \node [above=of #1_\the\mycount, yshift=#8*7cm] {$\linecontent$};
        \fi
        \advance\mycount by 1
    \repeat
    \closein\myfile
}

\newcommand{\addfailrow}[9]{
    \edef\temp{#9}
    
    \ifdefempty{\temp}{
        \node (#1_0) {\includegraphics[trim={#3cm, #4cm, #5cm, #6cm}, clip, width=7cm]{figures/raw_failure_split/#2_0.pdf}};
    }{
        \node (#1_0) [below=of #9_0, yshift=-0.9cm] {\includegraphics[trim={#3cm, #4cm, #5cm, #6cm}, clip, width=7cm]{figures/raw_failure_split/#2_0.pdf}};
    }

    \node (#1_1) [right=of #1_0, xshift=#7*7cm] {\includegraphics[trim={#3cm, #4cm, #5cm, #6cm}, clip, width=7cm]{figures/raw_failure_split/#2_1.pdf}};
    \node (#1_2) [right=of #1_1, xshift=#7*7cm] {\includegraphics[trim={#3cm, #4cm, #5cm, #6cm}, clip, width=7cm]{figures/raw_failure_split/#2_2.pdf}};
    \node (#1_3) [right=of #1_2, xshift=#7*7+0.6cm] {\includegraphics[trim={#3cm, #4cm, #5cm, #6cm}, clip, width=7cm]{figures/raw_failure_split/#2_remo.pdf}};

    \node (#1_text) at ([xshift=-0.5cm]#1_0.west) [anchor=south, rotate=90] {#8};
    
    \ifdefempty{\temp}{
        \draw[thick, gray] ([xshift=0.3cm]#1_2.south east) -- ([xshift=0.3cm, yshift=0.5cm]#1_2.north east);
    }{
        \draw[thick, gray] ([xshift=0.3cm]#1_2.south east) -- ([xshift=0.3cm]#9_2.south east);
    }
    
    \newcount\mycount
    \mycount=0

    \openin\myfile=figures/raw_failure_split/#2.txt
    \loop
        \unless\ifeof\myfile
        \read\myfile to \linecontent
        \ifnum\mycount<4
            \node [above=of #1_\the\mycount, yshift=0.01*7cm] {$\linecontent$};
        \fi
        \advance\mycount by 1
    \repeat
    \closein\myfile
}

\definecolor{wacvblue}{rgb}{0.21,0.49,0.74}
\usepackage[pagebackref,breaklinks,colorlinks,allcolors=wacvblue]{hyperref}

\def\confName{WACV}
\def\confYear{2026}

\begin{document}
\title{\paperTitle}
\author{\authorBlock}
\maketitle

\begin{abstract}
Future trajectories of neighboring traffic agents have a significant influence on the path planning and decision-making of autonomous vehicles.
While trajectory forecasting is a well-studied field, research mainly focuses on snapshot-based prediction, where each scenario is treated independently of its global temporal context.
However, real-world autonomous driving systems need to operate in a continuous setting, requiring real-time processing of data streams with low latency and consistent predictions over successive timesteps.
We leverage this continuous setting to propose a lightweight yet highly accurate streaming-based trajectory forecasting approach.
We integrate valuable information from previous predictions with a novel endpoint-aware modeling scheme.
Our temporal context propagation uses the trajectory endpoints of the previous forecasts as anchors to extract targeted scenario context encodings.
Our approach efficiently guides its scene encoder to extract highly relevant context information without needing refinement iterations or segment-wise decoding.
Our experiments highlight that our approach effectively relays information across consecutive timesteps. 
Unlike methods using multi-stage refinement processing, our approach significantly reduces inference latency, making it well-suited for real-world deployment.
We achieve state-of-the-art streaming trajectory prediction results on the Argoverse~2 multi-agent and single-agent benchmarks, while requiring substantially fewer resources.
\href{https://a-pru.github.io/seam/}{Project Page.}
\end{abstract}

\section{Introduction}
\label{sec:intro}

Trajectory prediction, \ie, forecasting the future movements of nearby traffic participants, is a key component of the autonomy stack in self-driving vehicles.
Trajectory prediction is essential for efficient and reliable navigation, as it enables autonomous vehicles to anticipate the behavior of surrounding agents and plan maneuvers proactively in response to predicted motions.
The latency of the trajectory prediction model directly contributes to the total delay between sensor observations and vehicle control actions as it is typically performed between the perception and motion planning modules.
Thus, for practical deployment in autonomous vehicles, trajectory prediction models must produce highly accurate predictions with minimal latency.

\begin{figure}[tp]
    \centering
    \includegraphics[trim={0cm, 0cm, 0cm, 0cm}, clip, width=0.93\linewidth]{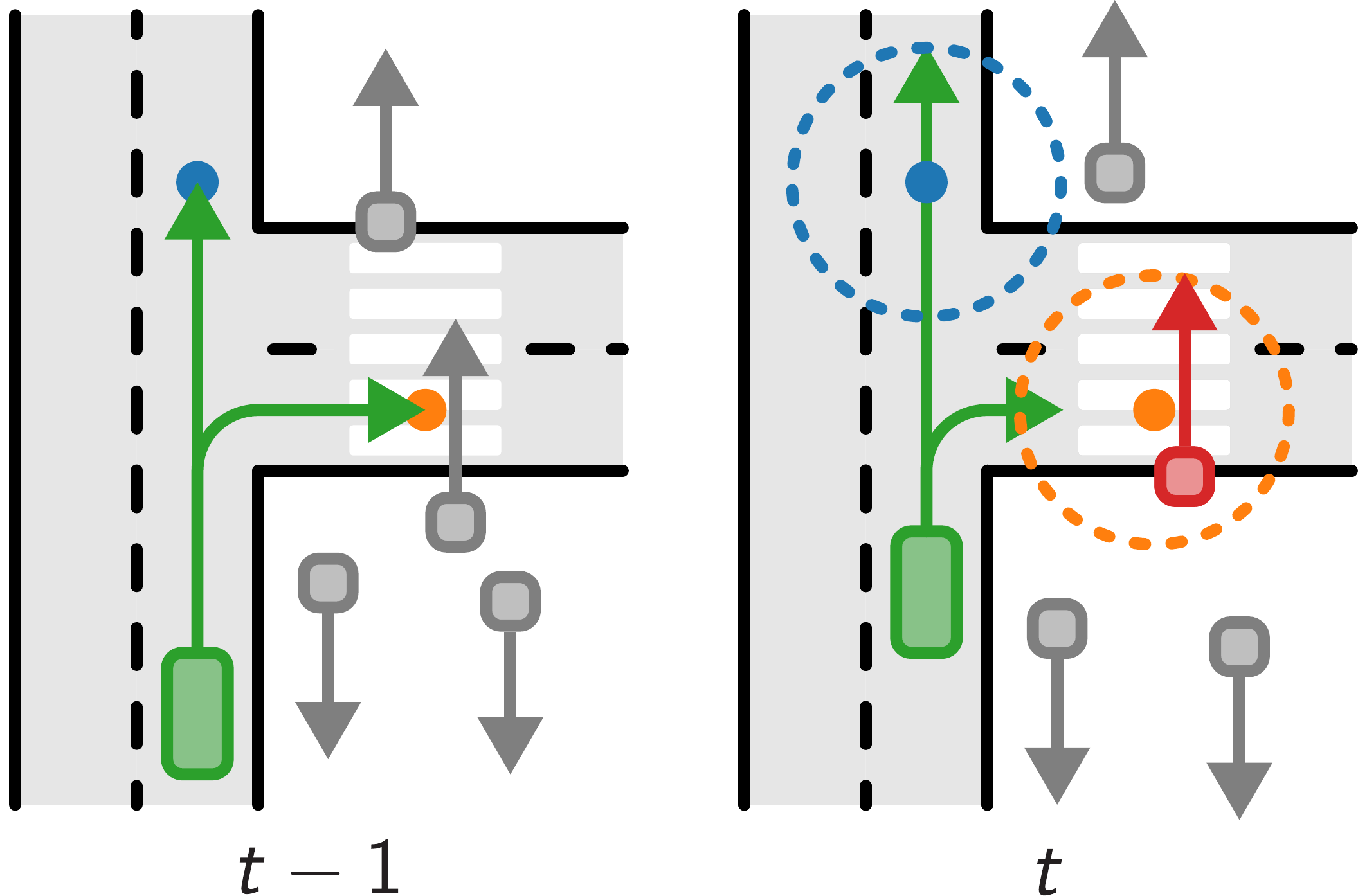}
    \vspace{-0.1cm}
    \caption{
    Traffic scenes are constantly changing, which is often neglected in trajectory prediction methods operating on snapshots rather than in a continuous setting.
    For example, for the trajectory prediction of the turning process of the focal agent (green), a possible interaction with a pedestrian (red) near the endpoint of the previous estimation (at time $t-1$) is important context that needs to be considered in the current prediction ($t$).
    We model this temporal context propagation with a novel endpoint streaming mechanism, achieving accurate predictions at minimal latencies.}
    \label{fig:teaser}
    \vspace{-0.35cm}
\end{figure}

One of the key challenges in trajectory forecasting is identifying the most influential map elements and the most relevant neighboring agents for motion prediction, especially in complex scenarios with long prediction horizons, often exceeding 100 meters.
While the initial agent position provides a strong prior for relevant context elements, scene elements along the future path play a crucial role, as illustrated in \Cref{fig:teaser}.
Recent trajectory prediction approaches commonly use refinement schemes~\cite{shi2022motion, zhou2023query, zhou2024smartrefine} to incorporate this knowledge: Initial coarse future trajectories are iteratively refined to enhance prediction accuracy.
For instance, SmartRefine~\cite{zhou2024smartrefine} proposes an adaptive multi-stage refinement scheme with scene context re-encoding to improve predictions of the strong baseline model QCNet~\cite{zhou2023query}.
Methods that use such refinement iterations achieve highly accurate results on large-scale benchmarks like Argoverse~2~(AV2)~\cite{wilson2021argoverse}.
However, they impose high computational cost and latency.
This limits their applicability in real-world autonomous systems, where fast and efficient decision-making is critical and methods with higher latency can lead to less reliable results due to outdated sensor observations.

In addition, current research in trajectory prediction focuses on processing and evaluating snapshots of individual scenarios in benchmark datasets~\cite{wilson2021argoverse, ettinger2021large}.
This snapshot-based treatment is in sharp contrast to the application in real autonomous vehicles, where trajectory prediction systems operate in a continuous environment.
In this real-world scenario, additional constraints must be considered.
In particular, the consistency across successive predictions, which is essential for integration into downstream decision-making systems.
On the other hand, continuous streaming-based processing also offers advantages, such as using contextual information across multiple prediction frames to improve the accuracy of future motion estimates.
Recently, pioneering streaming-based approaches~\cite{pang2023streaming, song2024realmotion} have emerged, laying the foundation for addressing the challenges of continuous trajectory prediction.
RealMotion~\cite{song2024realmotion} demonstrates that a simple backbone architecture combined with streaming mechanisms achieve promising results on the AV2 dataset~\cite{wilson2021argoverse}.
However, their approach does not use all available temporal information to encode the scene context, for example by not taking into account past predictions.

To advance towards consistent and efficient continuous real-world motion forecasting, we introduce a novel streaming-based trajectory prediction approach that uses temporal information for targeted scene context extraction.
In particular, we propose a \underline{s}treaming \underline{e}ndpoint-\underline{a}ware \underline{m}odeling~(\mn) architecture to effectively incorporate information across consecutive frames.
By leveraging prediction endpoints from the previous prediction frame as prior information, we directly integrate target region information into our modeling without requiring resource-intensive refinement iterations.
In-depth evaluations on AV2~\cite{wilson2021argoverse}, both single and multi-agent benchmarks, show that our new streaming mechanism provides significant advantages over previously proposed streaming techniques~\cite{song2024realmotion}.
We demonstrate that our approach outperforms more complex and resource-demanding models, while also addressing real-world considerations like reducing fluctuations across multiple predictions for the same agent.
It combines high accuracy with minimal inference latency, achieving state-of-the-art results for streaming-based trajectory prediction on the AV2 benchmarks.
\noindent In summary, our main contributions include:
\begin{itemize}
    \item We propose a new trajectory prediction model, effectively leveraging the consecutive nature of scenarios for target-centric context encoding without any refinement steps.
    \item Despite achieving minimal inference latency, we demonstrate that our model outputs excellent results on complex trajectory prediction tasks and sets a new state-of-the-art in streaming processing on the AV2 multi-agent benchmark.
    \item Our ablation study highlights the effectiveness of our endpoint-aware modeling in propagating information from the previous prediction timestep.
\end{itemize}

\section{Related Work}
\label{sec:related}
Trajectory prediction methods typically take agent states and map information as input.
Map data is predominantly modeled as vectorized data~\cite{gao2020vectornet, varadarajan2022multipath} and encoded by PointNet-like architectures~\cite{qi2017pointnet}, while agent time series information is commonly processed by recurrent architectures~\cite{mercat2020multi, varadarajan2022multipath}, attention-based modules~\cite{cheng2023forecast, lan2023sept, prutsch24efficient, song2024realmotion} or state-space models~\cite{zhang2024demo}.
To capture relationships between agents and map elements, recent methods utilize graph neural networks~\cite{gao2020vectornet, liang2020learning, zeng2021lanercnn, jia2023hdgt, cui2023gorela, tang2024hpnet} or attention-based encoders~\cite{liu2021multimodal, ngiam2021scene, shi2022motion, nayakanti2023wayformer, cheng2023forecast, zhang2023hptr, gan2024mgtr}.
Trajectory decoding is performed using either multilayer perceptrons~(MLPs)~\cite{cheng2023forecast, song2024realmotion} or more advanced cross-attention-based designs~\cite{shi2022motion, shi2024mtr++, zhang2024demo}.

\subsection{Refinement Modules for Trajectory Prediction}
Iterative refinement is a common strategy in state-of-the-art trajectory prediction approaches~\cite{shi2022motion, zhou2023query, wang2023ganet, gan2024mgtr, zhou2024smartrefine}.
Long-horizon benchmarks like Argoverse~2~\cite{wilson2021argoverse} and WOMD~\cite{ettinger2021large} involve predictions spanning over 100 meters, resulting in
a large number of scene context elements.
While the initial agent position helps to identify the relevant context, detecting critical elements at long distances remains challenging based solely on the given past motion of the agent.
To address this, multiple refinement steps can be applied to progressively extract scene context elements that have a strong influence on future agent behavior~\cite{shi2022motion, zhou2024smartrefine}.
Initial predictions provide strong guidance for identifying key scene elements which can be used to iteratively refine the trajectory hypotheses.

MTR~\cite{shi2022motion} and its follow-up works~\cite{shi2024mtr++, gan2024mgtr, demmler2025dynamic} iteratively refine trajectory predictions, using dynamic intention queries for aggregating scene context during decoding.
R-Pred~\cite{choi2023r} proposes a two-stage forecasting approach where initial predictions define scene context tubes which are then utilized by a refinement network to generate improved predictions.
QCNet~\cite{zhou2023query} employs a two-stage decoding process, where initial predictions are used as priors in a refinement stage by incorporating them into the mode query of a cross-attention decoder.
Additionally, at each inference step, it divides future trajectory prediction into shorter temporal segments, which are predicted recurrently.
SmartRefine~\cite{zhou2024smartrefine} proposes an adaptive refinement iteration strategy, which is demonstrated for QCNet.
At each inference step, an adaptive number of trajectory decoding iterations are performed to progressively refine predictions based on updated scene encoding.
Recently, DeMo~\cite{zhang2024demo} introduces a decoupled query strategy, first modeling motion state consistency and multi-modal intents separately before decoding the final trajectory output.

Overall, methods like DeMo~\cite{zhang2024demo} and SmartRefine~\cite{zhou2024smartrefine} provide strong evidence that multi-stage decoding improves trajectory prediction quality.
However, the gains in accuracy are often only small, especially considering the significantly higher computing costs required for executing a multi-stage refinement scheme at each prediction time step.
This diminishing return underscores the need for novel strategies to balance computational cost and accuracy, when designing trajectory prediction models for real-world applications.

\subsection{Streaming Trajectory Prediction}
Current trajectory prediction research primarily focuses on snapshot-based prediction, where individual scenarios are processed without considering global temporal context across evolving scenes.
In practice, however, autonomous systems operate in a continuous environment and process data in a streaming manner, introducing additional requirements like temporal consistency, as well as opportunities for improvement by leveraging additional temporal information.

Pang \etal~\cite{pang2023streaming}~present an early contribution to motion forecasting in continuous, streaming scenarios.
Their predictive streamer extends snapshot-based trajectory prediction models with a custom occlusion reasoning module and a differentiable filter to enhance temporal consistency.
RealMotion~\cite{song2024realmotion} shows that incorporating context and predictions from previous frames enables accurate results even with a simple backbone~\cite{cheng2023forecast}.
They introduce a context stream applying cross-attention on ego-motion-aligned context from current and previous frames, and a trajectory relay mechanism that refines current predictions via offsets from embedded past predictions.
DeMo~\cite{zhang2024demo} also integrates both mechanisms into their approach.
HPNet~\cite{tang2024hpnet} includes the exploration of a basic historical prediction attention module on the comparatively simple AV1 dataset~\cite{chang2019argoverse}.
QCNet~\cite{zhou2023query} explores a key-value cache for past observations, touching on concepts related to streaming processing. 
However, it does not consider past predictions or ensure alignment across prediction steps, and its compute-intensive decoder poses significant challenges for real-time deployment.

Overall, existing streaming-based methods incorporate mechanisms to maintain the consistency of successive predictions~\cite{pang2023streaming, song2024realmotion} and leverage the propagation of agent-centric scene context across timesteps~\cite{song2024realmotion} to address the continuous nature of real-world autonomous driving.
However, they do not explore the use of the previously predicted trajectories as prior knowledge for context encoding.
Given the importance of these regions for extracting relevant contextual features~\cite{shi2022motion, zhou2024smartrefine}, we propose an efficient strategy that includes this information in a single decoding step, thereby avoiding the high computational overhead of refinement modules.

\begin{figure*}[tp]
    \centering
    \includegraphics[trim={0cm, 0cm, 0cm, 0cm}, clip, width=0.99\linewidth]{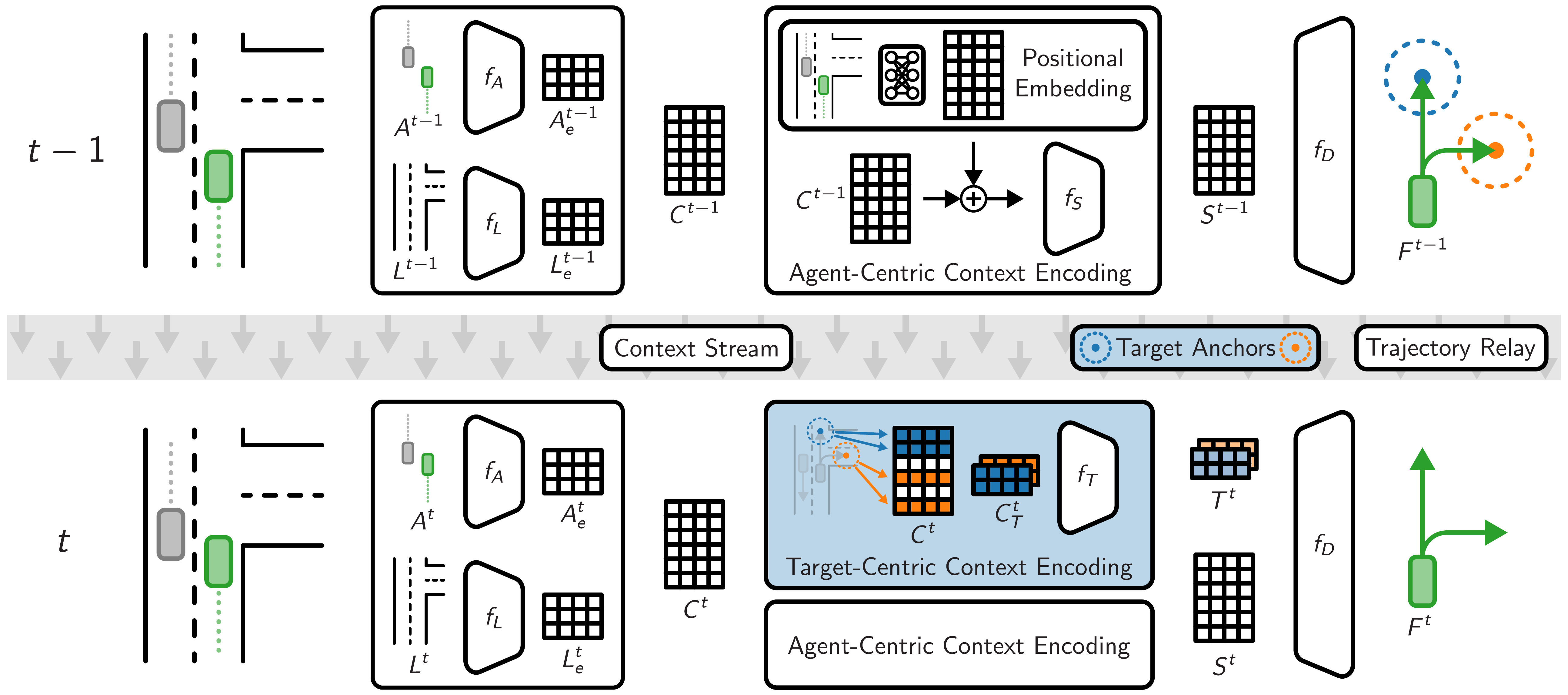}
    \vspace{-0.2cm}
    \caption{Overview of our streaming-based trajectory prediction architecture~\mn. We assume that for frame $t-1$ no previous prediction exists for our focal agent. Thus, only a standard trajectory prediction model pass is executed. In the next frame $t$, we incorporate the endpoints from the previous predictions to aggregate target-centric context information. We encode them using a second encoder path and provide it as additional input to our novel dual-context decoder.}
    \label{fig:arch}
    \vspace{-0.35cm}
\end{figure*}

\section{Streaming Trajectory Prediction Using Endpoint-Aware Modeling}
\label{sec:method}
We propose SEAM, an accurate yet lightweight trajectory prediction method based on a streaming-based processing paradigm.
Unlike refinement-based approaches~\cite{shi2022motion, zhou2024smartrefine}, which use multi-stage decoding to iteratively refine predictions by aggregating relevant scene context, SEAM introduces a novel mechanism that propagates predictions across timesteps to extract endpoint-centered target region features.
This design enables SEAM to achieve high accuracy with low latency by avoiding the computational overhead of refinement iterations.
Furthermore, by leveraging past predictions to identify potential target regions, SEAM addresses a key limitation of prior streaming-based methods~\cite{pang2023streaming, song2024realmotion}, which ignore this strong temporal prior.
Our approach aligns well with the continuous nature of real-world autonomous driving: as future prediction timesteps unfold, interactions with other agents may evolve and must be dynamically considered.
For example, recall \Cref{fig:teaser}: when predicting a turn, a potentially crossing pedestrian on the sidewalk must be considered.
As illustrated in overview \Cref{fig:arch}, we leverage past prediction endpoints to guide the model towards relevant regions, allowing us to achieve highly accurate future trajectories in a single decoding step, eliminating the need for computationally expensive refinement iterations.

The core component of our approach is the streaming-based dual-region encoding mechanism that uses the endpoints from past predictions to enhance scene context aggregation. 
In addition to conventional agent-centric context encoding, which is centered at the current position of the focal agent, we introduce a second set of context features:
We extract \emph{target-centric} features by retrieving scene elements closely around the endpoints of the previously predicted trajectory as these regions are likely to have high relevance for the current frame's prediction.
We encode them using a shallow target encoder with the trajectory endpoints as coordinate system origins.
During decoding, we execute attention to both agent-centric and target-centric features.

We describe the baseline encoding in \Cref{sec:alenc,sec:accenc}.
Next, we introduce our \emph{endpoint-aware modeling (EAM)}, which includes the novel target-centric feature encoding (\Cref{sec:tcce}) and its integration into the proposed dual-context decoder (\Cref{sec:dwdca}).
\Cref{sec:crtr} outlines the integration of baseline streaming mechanisms~\cite{song2024realmotion}. We present our extension to the multi-agent setting in \Cref{sec:ext-multiagent}.

\noindent \textbf{Input Representation:}
\quad Trajectory prediction refers to forecasting future trajectories $F$ given historical agent states $A$ and lane data $L$.
We follow the common input representation of state-of-the-art work~\cite{cheng2023forecast, prutsch24efficient, song2024realmotion, zhang2024demo} and initially sample all agents and map elements within a fixed radius around the focal agent. 
A tensor $A\in \mathbb{R}^{N_a \times T_h \times D_a}$ represents the historical trajectories, where $N_a$ is the number of agents, $T_h$  represents the number of historical timesteps and $D_a$ the feature dimension of the motion states, \ie positions and velocity.
Each lane is represented by sampling $P_l$ points along its centerline.
We define lane data as a tensor $L\in \mathbb{R}^{N_l \times P_l \times D_l}$, where $N_l$ denotes the number of lane segments and $D_l$ represents the feature dimension, \ie the $xy$-coordinates of the centerline.
To facilitate efficient encoding of historical movements and lane data, we first transform each element in a local coordinate system~\cite{cheng2023forecast}.
We normalize the historical trajectories \wrt their initial pose while coordinates of lane centerlines are given \wrt their center position.
This enables the model to learn motion patterns and lane segment shapes independently of global positions.

\noindent \textbf{Output Representation:}
\quad Our model predicts for a focal agent the multi-modal future trajectory hypotheses \mbox{$F \in \mathbb{R}^{ K \times T_f \times 2}$}, where $T_f$ denotes the number of future timesteps and $K$ the number of motion modes, along with associated probability scores $P \in \mathbb{R}^{K}$.
We represent future trajectories in $xy$-coordinates.
Thus, trajectory prediction can be formulated as $(F, P) = f(A, L)$, where $f$~is our trajectory prediction model.
More specifically, our model consists of an encoder $f_E$ generating the scene context $S = f_E(A, L)$ and a trajectory decoder $f_D$ using the scene context $S \in \mathbb{R}^{N_a+N_l \times D}$ to predict trajectories $F$ and probability scores~$P$ as $(F, P) = f_D(S)$.\\
\noindent \textbf{Streaming Processing:}
\quad To align with real-world autonomous driving applications, we adopt a streaming processing scheme to predict the future trajectories at the current timestep $t$.
We leverage previously encoded context information $S^{t-1}$ and past trajectory predictions $F^{t-1}$ to generate highly accurate predictions and ensure temporally consistent trajectories.
Consequently, our trajectory prediction at time $t$ can be formulated as $(F^t, P^t) = f(A^{t}, L^{t}, S^{t-1}, F^{t-1})$.

\subsection{Agent and Lane Encoding}
\label{sec:alenc}
We follow~\cite{lan2023sept, prutsch24efficient} to encode the historical agent movement by projecting the agent data $A^t$ into the $D$-dimensional feature space $\mathbb{R}^{N_a \times T_h \times D}$ using a linear layer.
Next, we apply multi-head self-attention~\cite{vaswani2017attention} along the temporal dimension to capture the agent motion dynamics and use a pooling operation to aggregate temporal information.
This yields an agent representation $A^t_e \in \mathbb{R}^{N_a \times D}$ where the movement of each agent is encoded in a feature vector $\mathbb{R}^{1 \times D}$.
To encode the lane data, we follow the default state-of-the-art approach~\cite{shi2022motion, shi2024mtr++, cheng2023forecast, prutsch24efficient, song2024realmotion, zhang2024demo} of using a mini-PointNet-like~\cite{qi2017pointnet} network.
This results in a lane context representation $L^t_e \in \mathbb{R}^{N_l \times D}$.
Thus, our encoders $f_A$ and $f_L$ generate agent feature tokens $A^t_e = f_A(A^t)$ and lane feature tokens $L^t_e = f_L(L^t)$ for further encoding.

\subsection{Agent-Centric Context Encoding}
\label{sec:accenc}
After encoding each scene context element individually, the relationships among all scene elements must be learned.
To achieve this, we concatenate the agent $A^t_e$ and lane $L^t_e$ tokens to form a scene context representation $C^t \in \mathbb{R}^{N_c \times D}$, where $N_c=N_a+N_l$.
To incorporate categorical information, like different agent and lane types, we add learnable type embeddings~\cite{cheng2023forecast} to the scene tokens.
A common way for encoding relations among map elements and agents is self-attention across scene tokens~\cite{cheng2023forecast, lan2023sept, prutsch24efficient, zhou2024smartrefine, zhang2024demo}.
To this end, a global positional encoding must be established, which is commonly done using a focal agent-centric coordinate system.
We encode the global poses using the position $(x, y)$ and the rotation $yaw$ of each local token coordinate system, \ie agent track origin and lane segment center.
We apply a sine and cosine transformation to the yaw and process the pose information $\mathbb{R}^{N_c \times 4}$ using a two-layer multilayer perceptron~(MLP).
Following, we add the positional embeddings to our context tokens $C^t \in \mathbb{R}^{N_c \times D}$ and apply multi-head attention in our scene encoder $f_S$.
This yields the encoded scene context $S^t=f_S(C^t)$ with~$S^t \in \mathbb{R}^{N_c \times D}$.

\subsection{Target-Centric Context Encoding}
\label{sec:tcce}
The agent-centric encoding implicitly guides the model to focus on the surroundings of the focal agent, where nearby context elements are highly relevant for predicting future motion.
However, as highlighted in \Cref{fig:teaser}, also the areas where the agent is heading to are highly important for accurate predictions.
Since we operate in a streaming-based setting, each agent that was visible in the previous frame has multi-modal past trajectory predictions, denoted as $F^{t-1} \in \mathbb{R}^{K \times T_f \times 2}$.
These previous predictions $F^{t-1}$ provide a strong prior for scenario regions that are also crucial for context extraction.
We use the endpoints of these trajectories as anchors to define $K$ target regions.
Next, we gather tokens from our scene context $C^t$ within a region-of-interest radius $r$ around each target anchor, forming $K$ individual target context token sets $C_T^t \in \mathbb{R}^{K \times N_c' \times D}$. 
In practice, $N_c'$ is significantly smaller than $N_c$ making our target encoding computationally less expensive than the main agent-centric encoding.

Following the same process as for the agent-centric scene context $S$, we encode each target-centric context $C_T^t$ individually in our target encoder $f_T$, obtaining $T^t \in \mathbb{R}^{K \times N_c' \times D}$.
To capture the relationship between the global coordinate root and each target anchor coordinate frame, we also add positional embeddings for the transformation to the target anchor of each region (see our supplementary for details).

\begin{figure}[tp]
    \centering
    \vspace{0.2cm}
    \includegraphics[trim={0cm, 0cm, 0cm, 0cm}, clip, width=\linewidth]{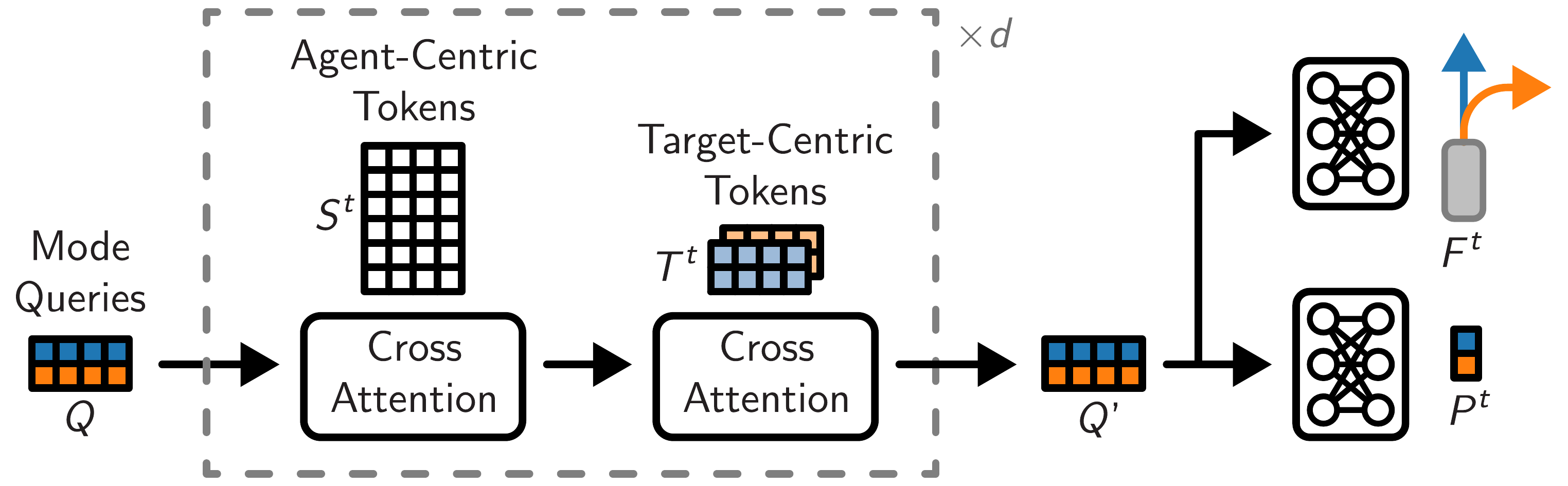}
    \vspace{-0.4cm}
    \caption{Architecture of our novel dual-context attention decoder, leveraging both agent-centric $S^t$ and target-centric $T^t$ features.}
    \label{fig:decoder}
    \vspace{-0.45cm}
\end{figure}

\subsection{Decoding with Dual-Context Attention}
\label{sec:dwdca}
Our decoder $f_D$ generates future trajectories $F^t$ and probability scores $P^t$ based on the agent-centric scene context $S^t$ and our novel target-centric scene context $T^t$, formulated as $(F^t, P^t) = f_D(Q, S^{t}, T^t)$.
We employ learnable queries $Q \in \mathbb{R}^{K \times D}$ to obtain multi-modal outputs and implement a DETR-like~\cite{carion2020end} decoding approach. 
\Cref{fig:decoder} illustrates the architecture of our dual-context attention decoder.

We iteratively execute cross-attention $d$-times to our scene context $S \in \mathbb{R}^{N_c \times D}$, and the target-specific tokens $T \in \mathbb{R}^{K \times N_c' \times D}$.
Thus, information from both feature sets is available in each decoder stage and integrated into our final output.
The agent-centric context $S$ provides local information, while our target context $T$ gives the model fine-grained information on long-horizon dependencies with other agents and map elements.
Each mode query only attends to one associated set of target tokens $T \in \mathbb{R}^{1 \times N_c' \times D}$.
Finally, we project the decoded queries $Q'$ to future trajectories $F^t \in \mathbb{R}^{K \times T_f \times 2}$ and probability scores $P^t \in \mathbb{R}^K$ using a dedicated two-layer MLP for each output.

\subsection{Context Referencing and Trajectory Relaying}
\label{sec:crtr}
In addition to our novel endpoint-aware modeling, we leverage context referencing and trajectory relaying~\cite{song2024realmotion} to model the streaming setting through information flow from the previous frame \( t\!-\!1 \).
We follow the setup of~\cite{song2024realmotion} for both mechanisms to enable fair comparison:
To account for focal agent motion between \( t\!-\!1 \) and $t$, we apply motion-aware layer normalization~(MLN)~\cite{wang2023exploring} to the scene context $S^{t-1}$ before integrating it into the current encoding $C^t$ via agent-to-scene and map-to-map cross-attention.
Trajectory relay employs a cross-attention-based interaction module to refine the current predictions $F^{t}$ using past predictions $F^{t-1}$.

\subsection{Extension to Multi-Agent Settings}
\label{sec:ext-multiagent}
To predict accurate trajectories for all agents $N_a$ in a scene, we first generate marginal predictions for each agent individually.
To this end, we treat each agent as focal agent once for our single-agent model, stack the input data and perform an inference pass with a batch size of $N_a$.
Next, we introduce a lightweight global consistency module, to fuse the marginal predictions to joint, globally consistent world predictions.

Our consistency module takes the mode features \mbox{$Q' \in \mathbb{R}^{N_a \times K \times D}$} for each agent and their global positions $R \in \mathbb{R}^{N_a \times 4}$ as input.
A shallow MLP encodes the global positions into positional embeddings.
We then apply self-attention across all modes per agent (over $K$) as well as self-attention across all agents per mode (over $N_a$) to capture intra- and inter-agent relationships.
Following our single agent decoder design, we decode the final world trajectory predictions using two shallow MLPs.
As output, we obtain $F_w \in \mathbb{R}^{K \times N_a \times T_f \times 2}$ world predictions with associated confidence scores $P_w \in \mathbb{R}^K$.

\section{Experiments}
\label{sec:experiments}
We present experimental results on the single-agent and multi-agent benchmarks of the Argoverse 2 (AV2)~\cite{wilson2021argoverse} dataset, along with a latency analysis for both settings to assess the practical applicability of our approach.
Additionally, we provide an ablation study for our endpoint-aware modeling on both AV2 and Argoverse~(AV1)~\cite{chang2019argoverse}, with further results included in the supplementary material.

\begin{table*}[t]
\small
\centering
\begin{tabular}{lc|ccccccc}
\toprule
Method & Streaming & minADE$_1$ & minFDE$_1$ & MR$_6$ & minADE$_6$ & minFDE$_6$ & brier-minFDE$_{6}$ \\ 
\midrule
SIMPL~\cite{zhang2024simpl}                       & \xmark & 2.03 & 5.50 & 0.19 & 0.72 & 1.43 & 2.05 \\
HPTR~\cite{zhang2023hptr}                         & \xmark & 1.84 & 4.61 & 0.19 & 0.73 & 1.43 & 2.03 \\
Forecast-MAE~\cite{cheng2023forecast}             & \xmark & 1.74 & 4.36 & 0.17 & 0.71 & 1.39 & 2.03 \\
MTR~\cite{shi2022motion}                          & \xmark & 1.74 & 4.39 & 0.15 & 0.73 & 1.44 & 1.98 \\
EMP-D~\cite{prutsch24efficient}                   & \xmark & 1.75 & 4.35 & 0.17 & 0.71 & 1.37 & 1.98 \\ 
GANet~\cite{wang2023ganet}                        & \xmark & 1.77 & 4.48 & 0.17 & 0.72 & 1.34 & 1.96 \\
ProIn~\cite{dong2024proin}                        & \xmark & 1.77 & 4.50 & 0.18 & 0.73 & 1.35 & 1.93 \\
RealMotion~\cite{song2024realmotion} *            & \cmark & 1.65 & 4.12 & 0.15 & 0.67 & 1.29 & 1.93 \\
QCNet~\cite{zhou2023query}                        & \xmark & 1.69 & 4.30 & 0.16 & 0.65 & 1.29 & 1.91 \\
RealMotion~\cite{song2024realmotion} (Train/Val)* & \cmark &\sval{1.59} & \sval{3.93} & 0.15 & 0.66 & 1.24 & 1.89 \\
Tamba~\cite{huang2025trajectory}                  & \xmark & 1.66 & 4.24 & 0.17 & 0.64 & 1.24 & 1.89 \\
ProphNet~\cite{wang2023prophnet}                  & \xmark & 1.80 & 4.74 & 0.18 & 0.68 & 1.33 & 1.88 \\
MTR++~\cite{shi2024mtr++}                         & \xmark & 1.64 & 4.12 & \sval{0.14} & 0.71 & 1.37 & 1.88 \\
DyMap~\cite{fan2025bidirectional}                 & \xmark & -    & -    & - & 0.71 & 1.29 & 1.87 \\
SmartRefine~\cite{zhou2024smartrefine}            & \xmark & 1.65 & 4.17 & 0.15 & \sval{0.63} & \sval{1.23} & \sval{1.86} \\
DeMo~\cite{zhang2024demo}                         & \cmark & \bval{1.49} & \bval{3.74} &\bval{0.13} & \bval{0.61} & \bval{1.17} & \bval{1.84} \\
\rowcolor{rcol}\mn~(Ours)                         & \cmark & 1.60 & 3.95 & 0.15 & 0.66 & 1.24 & \bval{1.84} \\ %
\bottomrule
\end{tabular}
\vspace{-0.1cm}
\caption{
Single-agent trajectory prediction results on the Argoverse~2 test set.
For all metrics lower indicate better, table sorted in descending order by brier-minFDE$_{6}$.
We report results for published works on the official leaderboard without model ensembling.
The streaming approach of Pang \etal~\cite{pang2023streaming} is not included (their custom dataset prevents direct comparison).
*For RealMotion~\cite{song2024realmotion} we report the results for their official checkpoint, as well as the results from their paper, which is trained on the train and validation set.
}
\vspace{-0.5cm}
\label{tab:res_av2_test}
\end{table*}

\subsection{Experimental Settings}
\noindent \textbf{Dataset:}
\quad We evaluate our model on the Argoverse~2~(AV2)~\cite{wilson2021argoverse} trajectory prediction dataset.
It has a sampling rate of 10 Hz and contains 199,908 training, 24,988 validation, and 24,984 test scenarios collected in six U.S. cities.
In each scenario the first \seconds{5} serve as historical context and the goal is to predict the trajectories for the subsequent \seconds{6}.
The long historical context makes AV2 ideal for evaluation in streaming processing, as it allows a reasonable sliding-window processing (see below).
In contrast, WOMD~\cite{ettinger2021large} provides only \seconds{1} of history and nuScenes~\cite{caesar2020nuscenes} offers \seconds{2} but at only 2\,Hz.
Further reducing this few historical data points to multiple windows limits the agent encoder’s ability to learn motion patterns.
Both are also not considered for streaming processing by related work~\cite{pang2023streaming, song2024realmotion, zhang2024demo}.
We evaluate on the single-agent (predict trajectories for one agent per scenario) and multi-agent (predict globally consistent trajectories for multiple agents per scenario) benchmarks of AV2.
Additionally, we provide an ablation study on Argoverse~1~(AV1)~\cite{chang2019argoverse} which features \seconds{2} history and considers a future of \seconds{3}.

\noindent \textbf{Metrics:}
\quad We evaluate our approach using the standard AV2~\cite{wilson2021argoverse} benchmark metrics.
For the single-agent case, we use the miss rate (MR$_k$), minimum average displacement error (minADE$_k$), minimum final displacement error (minFDE$_k$), and Brier minimum final displacement error (brier-minFDE$_k$).
Following the official leaderboard, we evaluate using the top $k \in \{1, 6\}$ scoring trajectories.
For the multi-agent case we utilize the actorMR$_k$, avgMinADE$_k$, avgMinFDE$_k$ and avgBrierMinFDE$_k$ which account for the mean errors across all actors in each world.

\noindent \textbf{Streaming Processing:}
\quad To enable streaming processing using existing benchmarks, we follow~\cite{song2024realmotion, zhang2024demo} by limiting the historical model input $T_h$ to a fraction of the available history.
We then operate using a sliding window and conduct multiple model executions.
In particular, for AV2 we use a historic input of \seconds{h=3} and execute three predictions at $t \in \{3, 4, 5\}$s.
The final prediction is made at the same time point as in snapshot-based methods, allowing a fair comparison. 
We provide a detailed comparison of streaming and snapshot-based processing in our supplementary material.

\begin{table*}[t]
\small
\centering
\setlength{\tabcolsep}{6pt}
\begin{tabular}{lc|cccccc}
\toprule
Method & Streaming & avgMinADE$_1$ & avgMinFDE$_1$ & actorMR$_6$ & avgMinADE$_6$ & avgMinFDE$_6$ & avgBrierMinFDE$_6$ \\ 
\midrule
FJMP~\cite{rowe2023fjmp}              & \xmark & 1.52 & 4.00 & 0.23 & 0.81 & 1.89 & 2.59 \\
Forecast-MAE~\cite{cheng2023forecast} & \xmark & 1.30 & 3.33 & 0.19 & 0.69 & 1.55 & 2.24 \\
RealMotion~\cite{song2024realmotion}  & \cmark & 1.14 & 2.97 & 0.18 & 0.62 & 1.32 & 2.01 \\
DeMo~\cite{zhang2024demo}             & \cmark & 1.12 & 2.78 & \sval{0.16} & \sval{0.58} & 1.24 & 1.93 \\
\rowcolor{rcol}\mn~(Ours)             & \cmark & \sval{1.05} & \sval{2.57} & \sval{0.16} & 0.62 & \sval{1.21} & \sval{1.85} \\ %
QCNeXt~\cite{zhou2023qcnext}          & \xmark & \bval{1.03} & \bval{2.55} & \bval{0.14} & \bval{0.54} & \bval{1.13} & \bval{1.79} \\
\bottomrule
\end{tabular}
\vspace{-0.1cm}
\caption{
Multi-agent trajectory prediction results on the Argoverse~2 test set.
For all metrics lower values indicate better performance, table sorted in descending order by avgBrierMinFDE$_6$.
For all methods we report results without model ensembling.
}
\vspace{-0.2cm}
\label{tab:res_av2_multi}
\end{table*}

\subsection{Argoverse 2 Single-Agent Results}
Table~\ref{tab:res_av2_test} shows the results for the Argoverse~2 single-agent test set. 
Our SEAM, along with DeMo~\cite{zhang2024demo}, achieves the best score on the main metric (brier-minFDE$_6$), outperforming  recent work like Tamba~\cite{huang2025trajectory} and even more resource-intensive methods, such as QCNet~\cite{zhou2023query} and SmartRefine~\cite{zhou2024smartrefine}.
This strong performance without any refinement iterations highlights the effectiveness of our endpoint-aware modeling.
We also outperform RealMotion~\cite{song2024realmotion}, a prior streaming-based prediction method, while also achieving lower inference latency (\cref{sec:latency}).
DeMo~\cite{zhang2024demo} achieves slightly lower displacement errors since their decoupled decoding yields more diverse trajectories, but at a significantly higher latency (\cref{sec:latency}).
The small gap between minFDE$_6$ to brier-minFDE$_6$ (induced by the main metric's probability penalty) demonstrates our model’s strong confidence estimation, which is also crucial for downstream integration into the planning module.
This highlights that directly integrating target-centric features helps assess maneuver likelihood.

\subsection{Argoverse 2 Multi-Agent Results}
Previous methods, \eg~\cite{zhou2024smartrefine, huang2025trajectory}, often report only single-agent results.
However, the multi-agent benchmark is more relevant for practical systems: models deployed in real-world settings must be capable to predict trajectories for more than one agent in a scene at the same time.
The results in Table~\ref{tab:res_av2_multi} show that, in this setting, our \mn~outperforms previous streaming-based methods~\cite{song2024realmotion, zhang2024demo} in the main metrics.
The current best-performing method is the snapshot method QCNeXt~\cite{zhou2023qcnext}, an extension of QCNet~\cite{zhou2023query}.
However, the displacement errors differ only by a few centimeters, whereas its extensive use of recurrence and refinement iterations in decoding poses significant challenges for real-time deployment.
Since model latency directly impacts the delay between sensor observations and prediction output, models with computational overhead may lead to worse results in real-world scenarios due to delayed response.
For reference, an agent traveling at 50 km/h covers approximately $0.7$~m in 50~ms. 
Thus, marginally lower displacement errors at increased processing delay offer no benefit in practice.

\begin{table*}[t]
\small
\centering
\begin{tabular}{lc|cc|cc|cc|c| c}
\toprule
Method & Streaming & \multicolumn{2}{c|}{Latency ($B=1$)} & \multicolumn{2}{c|}{Latency ($B=32$)} & \multicolumn{2}{c|}{Latency ($B=64$)} & Model & AV2 Test Set \\ 
       &           &Offline & Online & Offline & Online & Offline & Online & Parameters &   brier-minFDE$_{6}$\\
\midrule
SmartRefine~\cite{zhou2024smartrefine}  & \xmark & \multicolumn{2}{c|}{$>$118~ms*} & \multicolumn{2}{c|}{$>$680~ms*} & \multicolumn{2}{c|}{-} & 8.0M & \sval{1.86}  \\
QCNet~\cite{zhou2023query}              & \xmark & \multicolumn{2}{c|}{\phantom{$>$}118~ms\phantom{$*$}} & \multicolumn{2}{c|}{\phantom{$>$}680~ms\phantom{*}} & \multicolumn{2}{c|}{-} & 7.7M & 1.91 \\
RealMotion~\cite{song2024realmotion}    & \cmark & \bval{\phantom{0}86~ms} & \bval{24~ms} & \sval{247~ms} & 88~ms & \sval{436~ms} & 150~ms & \bval{2.9M} & 1.93 \\
DeMo~\cite{zhang2024demo}               & \cmark & 139~ms & 39~ms & 275~ms & \sval{83~ms} & 466~ms & \sval{142~ms} & 5.9M & \bval{1.84} \\
\rowcolor{rcol}\mn~(Ours)               & \cmark & \phantom{0}\sval{91~ms} & \sval{28~ms} & \bval{140~ms} & \bval{50~ms} & \bval{246~ms} & \phantom{0}\bval{74~ms} & \sval{4.6M} & \bval{1.84} \\
\bottomrule
\end{tabular}
\vspace{-0.1cm}
\caption{
Latency analysis for predicting $B$ Argoverse~2 scenarios in the single-agent setting using a NVIDIA V100 GPU.
\emph{Online} (for streaming-based methods) is the time required to process a single sliding window; \emph{Offline} is the time required to process the full stream (to generate the benchmark predictions). 
For streaming methods the \emph{online} latency is relevant for practical applications. %
*SmartRefine does not provide an implementation for AV2---since its main results add refinement iterations to QCNet, latency is higher than for QCNet.
}
\vspace{-0.3cm}
\label{tab:latencyv100}
\end{table*}

\subsection{Latency Analysis}
\label{sec:latency}
\noindent \textbf{Argoverse~2 Single-Agent Setting:}
\quad 
Table~\ref{tab:latencyv100} compares latencies for predicting various batch sizes of AV2 single-agent scenarios. %
Our approach has favorable runtime compared to related work, especially since it scales well with larger batch sizes which are relevant in practice as the single-agent benchmark evaluates only one agent per scenario.
\mn~achieves minimal latency by using only a single decoder pass, unlike DeMo~\cite{zhang2024demo}, QCNet~\cite{zhou2023query} and SmartRefine~\cite{zhou2024smartrefine}, which employ multiple decoding stages.
Additionally, it predicts the complete future trajectory without splitting it into segments, as done in QCNet and SmartRefine.
In comparison to RealMotion~\cite{song2024realmotion} our more efficient decoder design leads to a latency advantage.
The fast processing speed makes our approach highly suitable for real-world deployment, as it combines streaming modeling with computational efficiency --- two key considerations for practical use.

\noindent \textbf{Argoverse~2 Multi-Agent Setting:}
\quad 
Related work on streaming-based processing (RealMotion~\cite{song2024realmotion} and DeMo~\cite{zhang2024demo}) does not provide details on their multi-agent implementations.
Moreover, no official implementation is available for QCNeXt~\cite{zhou2023qcnext}.
As a result, a direct latency comparison is not possible.
However, based on the nature of the task, we can reasonably assume that per-scenario latency in a multi-agent setting exceeds the single-agent latencies reported in Table~\ref{tab:latencyv100}.
For our model, we measure an average online latency of \emph{38~ms} per scenario in the multi-agent setting on the AV2 validation set using a single NVIDIA V100 GPU (batch size $B=1$).
Notably, this is even faster than the single-agent prediction latencies reported for DeMo and QCNet, highlighting the high efficiency of our approach.

\begin{figure*}[tp]
    \centering
    \resizebox{\linewidth}{!}{
        \begin{tikzpicture}[node distance=0mm, inner sep=0mm]
            \addresrow{0}{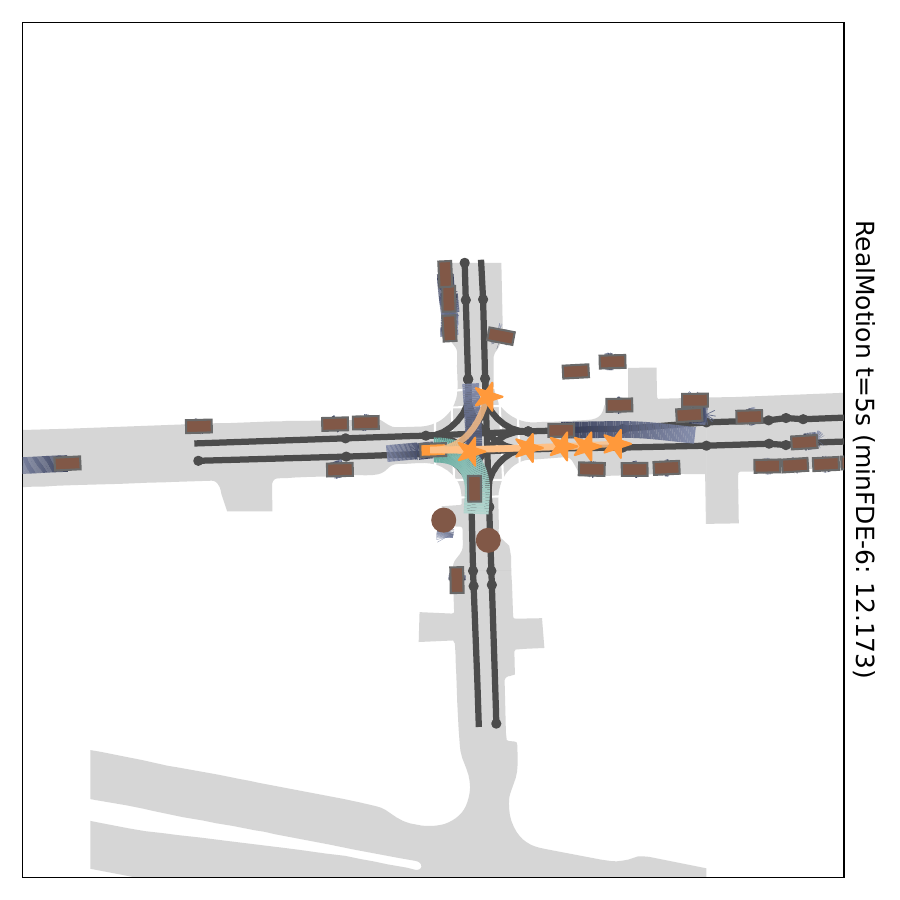}{5}{5.5}{3}{6.5}{0.025}{0.01}{}
            \addresrow{1}{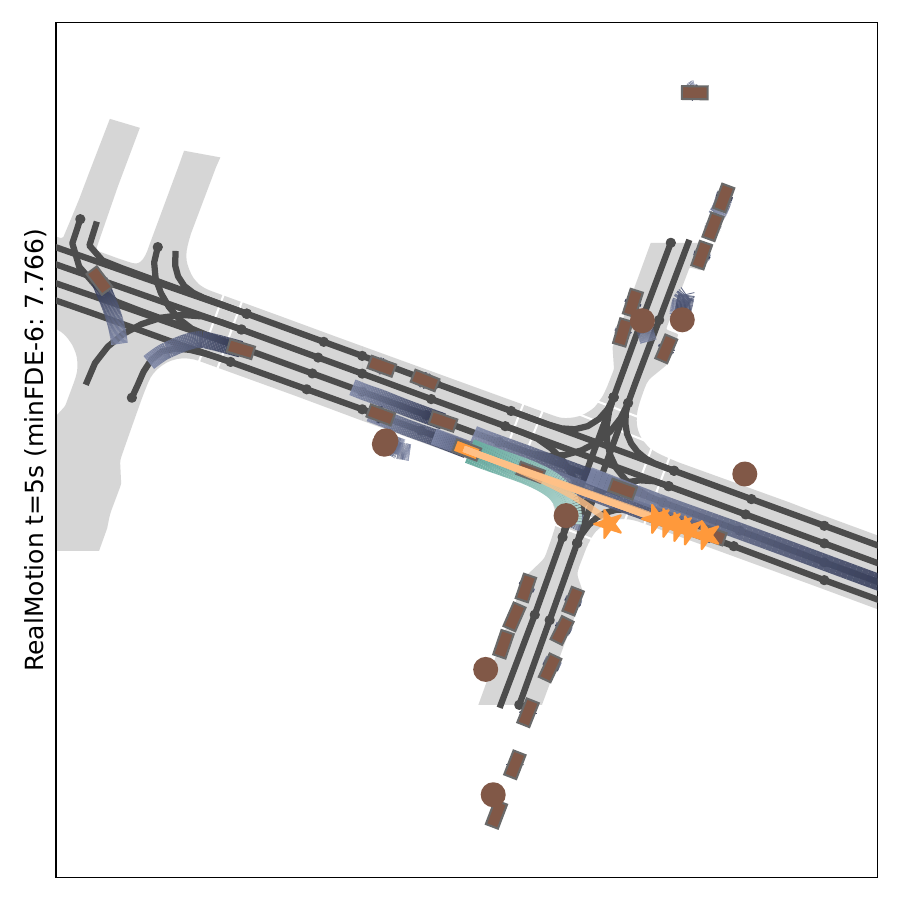}{4}{5.3}{1.8}{7}{0.025}{0.01}{0}
        \end{tikzpicture}
    }
    \vspace{-0.5cm}
    \caption{Qualitative results on two Argoverse~2 scenarios. We show the \textcolor[HTML]{ff9a3a}{\textbf{predictions}} of our streaming-based method at $t \in \{3, 4, 5\}$s.
    The visualizations also show \textcolor[rgb]{0.41, 0.67, 0.63}{\textbf{ground truth future}}, \textcolor[HTML]{384062}{\textbf{agent histories}}, and \textcolor[HTML]{815847}{\textbf{neighboring agents}}.
    The right column shows the final predictions at $t = 5$s for RealMotion~\cite{song2024realmotion} in the streaming-based setting.
    Top row: the focal agent is approaching an intersection where other traffic is currently passing by, making it difficult to identify possible movement.
    Bottom row: a pedestrian crosses the street at an intersection to the right.
    Our approach correctly predicts that the vehicle can either continue straight or turn right, either waiting before the crosswalk or proceeding directly if the pedestrian has already left the crosswalk.
    }
    \vspace{-0.4cm}
    \label{fig:results}
\end{figure*}

\begin{table}[t]
\centering
\setlength{\tabcolsep}{2pt}
\footnotesize{
\begin{tabular}{ccccc|ccc}
\toprule
Dataset & $h$ & Stream & CS+TR~\cite{song2024realmotion} & EAM & mADE$_6$ & mFDE$_6$ & b-mFDE$_6$ \\ 
\midrule
 & 2s & \xmark & \xmark & \xmark & 0.64 & 1.04 & 1.67 \\ %
AV1 & 1s & \cmark & \cmark & \xmark                 & 0.62 & 0.98 & 1.62 \\ %
Val & 1s & \cmark & \xmark & \cmark                 & 0.60 & 0.95 & 1.55 \\ %
 & \cellcolor{rcol}1s & \cellcolor{rcol}\cmark & \cellcolor{rcol}\cmark & \cellcolor{rcol}\cmark  & \cellcolor{rcol}\bval{0.59} & \cellcolor{rcol}\bval{0.93} & \cellcolor{rcol}\bval{1.52} \\ %
\midrule
& 5s & \xmark & \xmark & \xmark & 0.77 & 1.39 & 2.00 \\ %
AV2 & 3s & \cmark & \cmark & \xmark  & 0.75 & 1.29 & 1.93 \\ %
Val & 3s & \cmark & \xmark & \cmark  & {0.70} & {1.27} & {1.88} \\ %
 & \cellcolor{rcol}3s & \cellcolor{rcol}\cmark & \cellcolor{rcol}\cmark & \cellcolor{rcol}\cmark & \cellcolor{rcol}\bval{0.66} & \cellcolor{rcol}\bval{1.25} & \cellcolor{rcol}\bval{1.85} \\  %
\bottomrule
\end{tabular}
}
\vspace{-0.1cm}
\caption{
Comparison of different information forwarding mechanisms for streaming trajectory prediction.
We compare the integration of context stream~(CS) and trajectory relay~(TR) introduced by~\cite{song2024realmotion} to our endpoint-aware modeling~(EAM). The first line shows the traditional snapshot-based use.
We highlight \textbf{best} values.
}
\vspace{-0.5cm}
\label{tab:abl_relay}
\end{table}

\subsection{Ablation Studies}

\noindent \textbf{Information Streaming Mechanisms:}
\quad In Table~\ref{tab:abl_relay} we compare our endpoint-aware modeling (EAM), which includes the target-centric feature encoding and the dual-context decoder, to the context stream~({CS}) and trajectory relay~({TR}) streaming mechanisms introduced by previous work~\cite{song2024realmotion}.
The top row in each group presents the results of our backbone architecture operating in the traditional snapshot-based prediction setting without any streaming mechanism.
Comparing streaming-based processing using {CS} and {TR} to our approach highlights the effectiveness of our EAM in propagating information from the previous frame.
Our prediction endpoint-aware modeling achieves a larger improvement than the two streaming mechanisms proposed by~\cite{song2024realmotion}.
Leveraging endpoint information in the streaming setting proves to be highly beneficial for enhancing trajectory prediction accuracy.
Overall, combining our EAM with {CS} and {TR} leads to the best results.

\noindent \textbf{Target Context Radius:}
\quad 
We evaluate the influence of different target context radii $r$ in Table~\ref{tab:abl_radius}.
Performance saturates at a radius of 30 meters, as incorporating further context at larger radii does not provide relevant information for our endpoint-aware modeling.
A larger radius increases the number of target context tokens and thus latency, making a radius of 30 meters an optimal trade-off between performance and efficiency for practical applications.
Notably, our approach remains effective even with a smaller radius of 10 meters, as injecting previous endpoint positions into the decoder already provides a strong prior. 
Incorporating the endpoint location during decoding also guides cross-attention toward agent-centric features, further underscoring the importance of endpoint information in streaming processing.

\begin{table}[t]
\centering
\setlength{\tabcolsep}{6pt}
\footnotesize{
\begin{tabular}{cc|ccccccc}
\toprule
Dataset & $r$ & mADE$_6$ & mFDE$_6$ & b-mFDE$_6$ \\ 
\midrule
& 10m & 0.68 & 1.27 & 1.89      \\ %
AV2 & 15m & 0.67 & 1.26 & 1.87      \\ %
 Val & \cellcolor{rcol}30m & \cellcolor{rcol}\bval{0.66} & \cellcolor{rcol}\bval{1.25} & \cellcolor{rcol}\bval{1.85}  \\ %
& 45m & \bval{0.66} & \bval{1.25} & \bval{1.85}  \\ %
\bottomrule
\end{tabular}
}
\vspace{-0.1cm}
\caption{
Comparison of different radii $r$ to select context elements for our target-centric region.
We highlight \textbf{best} values.
}
\vspace{-0.5cm}
\label{tab:abl_radius}
\end{table}

\subsection{Qualitative Results}
\Cref{fig:results} shows predictions for two AV2 validation set scenarios, illustrating the evolving nature of streaming processing at $t \in \{3, 4, 5\}$s.
To demonstrate the benefits of our endpoint-aware modeling against previous work on streaming trajectory prediction, we also show the output of RealMotion~\cite{song2024realmotion} at $t=5$s.
Additional qualitative results, including failure cases, are available in the supplementary material.

\section{Conclusion}
\label{sec:conclusion}
We propose a novel endpoint-aware modeling for streaming-based trajectory prediction.
Our dual-context approach enables us to effectively incorporate the continuously evolving relevant scene context, eliminating the need for costly trajectory refinement iterations to achieve highly accurate predictions.
We significantly outperform inference speed of highly iterative schemes~\cite{shi2024mtr++, zhou2024smartrefine} and previous streaming based methods~\cite{zhang2024demo, song2024realmotion} for practical scenario sizes, while achieving competitive results on the Argoverse~2 single-agent dataset and setting new state-of-the-art for streaming-based trajectory prediction on the multi-agent benchmark.

{\small
\vspace{6pt}
\nbf{Acknowledgments}
This work was partially funded by Addsafety (923936), a COMET Project funded by BMIMI, BMWET as well as the co-financing federal province of Styria. The COMET programme is managed by the FFG.
}

{
    \small
    \bibliographystyle{ieeenat_fullname}
    \bibliography{_sections/11_references}

@String(PAMI = {IEEE Trans. Pattern Anal. Mach. Intell.})

@String(CVPR= {IEEE Conf. Comput. Vis. Pattern Recog.})

@String(ICCV= {Int. Conf. Comput. Vis.})

@String(ECCV= {Eur. Conf. Comput. Vis.})

@String(NIPS= {Adv. Neural Inform. Process. Syst.})

@String(ICLR = {Int. Conf. Learn. Represent.})

@String(TASE  = {IEEE Trans. Automation Science and Eng.})

@String(PAMI  = {IEEE TPAMI})

@String(CVPR  = {CVPR})

@String(ICCV  = {ICCV})

@String(ECCV  = {ECCV})

@String(NIPS  = {NeurIPS})

@String(ICLR  = {ICLR})

@String(IROS  = {IROS})

@String(ICRA  = {ICRA})

@String(iv  = {IEEE IV})

@article{hendrycks2016gaussian,
  title={{Gaussian Error Linear Units (GELUs)}},
  author={Hendrycks, Dan and Gimpel, Kevin},
  journal={arXiv},
  year={2016}
}

@inproceedings{vaswani2017attention,
  title={{Attention Is All You Need}},
  author={Vaswani, Ashish and Shazeer, Noam and Parmar, Niki and Uszkoreit, Jakob and Jones, Llion and Gomez, Aidan N and Kaiser, {\L}ukasz and Polosukhin, Illia},
  booktitle=NIPS,
  year={2017}
}

@inproceedings{tang2024hpnet,
  title={{HPNet: Dynamic Trajectory Forecasting with Historical Prediction Attention}},
  author={Tang, Xiaolong and Kan, Meina and Shan, Shiguang and Ji, Zhilong and Bai, Jinfeng and Chen, Xilin},
  booktitle=cvpr,
  year={2024}
}

@inproceedings{choi2023r,
  title={{R-Pred: Two-Stage Motion Prediction Via Tube-Query Attention-Based Trajectory Refinement}},
  author={Choi, Sehwan and Kim, Jungho and Yun, Junyong and Choi, Jun Won},
  booktitle=iccv,
  year={2023}
}

@article{ngiam2021scene,
  title={{SceneTransformer: A Unified Architecture for Predicting Multiple Agent Trajectories}},
  author={Ngiam, Jiquan and Caine, Benjamin and Vasudevan, Vijay and Zhang, Zhengdong and Chiang, Hao-Tien Lewis and Ling, Jeffrey and Roelofs, Rebecca and Bewley, Alex and Liu, Chenxi and Venugopal, Ashish and others},
  journal=iclr,
  year={2022}
}

@article{jia2023hdgt,
  title={{HDGT: Heterogeneous Driving Graph Transformer for Multi-Agent Trajectory Prediction via Scene Encoding}},
  author={Jia, Xiaosong and Wu, Penghao and Chen, Li and  Liu, Yu  and Li, Hongyang and Yan, Junchi},
  journal = PAMI,
  year = {2023},
}

@inproceedings{gao2020vectornet,
  title={{VectorNet: Encoding HD Maps and Agent Dynamics From Vectorized Representation}},
  author={Gao, Jiyang and Sun, Chen and Zhao, Hang and Shen, Yi and Anguelov, Dragomir and Li, Congcong and Schmid, Cordelia},
  booktitle=cvpr,
  year={2020}
}

@article{zhou2023qcnext,
  title={{QCNeXt: A Next-Generation Framework For Joint Multi-Agent Trajectory Prediction}},
  author={Zhou, Zikang and Wen, Zihao and Wang, Jianping and Li, Yung-Hui and Huang, Yu-Kai},
  journal={arXiv},
  year={2023}
}

@inproceedings{rowe2023fjmp,
  title={{FJMP: Factorized Joint Multi-Agent Motion Prediction over Learned Directed Acyclic Interaction Graphs}},
  author={Rowe, Luke and Ethier, Martin and Dykhne, Eli-Henry and Czarnecki, Krzysztof},
  booktitle=cvpr,
  year={2023}
}

@inproceedings{wang2023ganet,
  title={{GANet: Goal Area Network for Motion Forecasting}},
  author={Wang, Mingkun and Zhu, Xinge and Yu, Changqian and Li, Wei and Ma, Yuexin and Jin, Ruochun and Ren, Xiaoguang and Ren, Dongchun and Wang, Mingxu and Yang, Wenjing},
  booktitle=icra,
  year={2023},
}

@article{dong2024proin,
  title={{ProIn: Learning to Predict Trajectory Based on Progressive Interactions for Autonomous Driving}},
  author={Dong, Yinke and Yuan, Haifeng and Liu, Hongkun and Jing, Wei and Li, Fangzhen and Liu, Hongmin and Fan, Bin},
  journal={arXiv},
  year={2024}
}

@article{fan2025bidirectional,
  title={{Bidirectional Agent-Map Interaction Feature Learning Leveraged by Map-Related Tasks for Trajectory Prediction in Autonomous Driving}},
  author={Fan, Bin and Yuan, Haifeng and Dong, Yinke and Zhu, Zhengyu and Liu, Hongmin},
  journal=TASE,
  year={2025},
}

@article{shi2024mtr++,
  title={{MTR++: Multi-Agent Motion Prediction With Symmetric Scene Modeling and Guided Intention Querying}},
  author={Shi, Shaoshuai and Jiang, Li and Dai, Dengxin and Schiele, Bernt},
  journal=PAMI,
  year={2024},
  publisher={IEEE}
}

@inproceedings{wang2023exploring,
  title={{Exploring Object-Centric Temporal Modeling for Efficient Multi-View 3D
Object Detection}},
  author={Wang, Shihao and Liu, Yingfei and Wang, Tiancai and Li, Ying and Zhang, Xiangyu},
  booktitle=cvpr,
  year={2023}
}

@inproceedings{pang2023streaming,
  title={{Streaming Motion Forecasting for Autonomous Driving}},
  author={Pang, Ziqi and Ramanan, Deva and Li, Mengtian and Wang, Yu-Xiong},
  booktitle=iros,
  year={2023}
}

@inproceedings{zhou2023query,
  title={{Query-Centric Trajectory Prediction}},
  author={Zhou, Zikang and Wang, Jianping and Li, Yung-Hui and Huang, Yu-Kai},
  booktitle=cvpr,
  IGNOREpages={17863--17873},
  year={2023}
}

@inproceedings{prutsch24efficient,
  title = {{Efficient Motion Prediction: A Lightweight \& Accurate Trajectory Prediction Model With Fast Training and Inference Speed}},
  author = {Prutsch, Alexander and Bischof, Horst and Possegger, Horst},
  booktitle = iros,
  year = {2024}
}

@inproceedings{zhou2024smartrefine,
  title={{SmartRefine: A Scenario-Adaptive Refinement Framework for Efficient Motion Prediction}},
  author={Yang Zhou and Hao Shao and Letian Wang and Steven L. Waslander and Hongsheng Li and Yu Liu},
  booktitle = cvpr,
  year = {2024}
}

@inproceedings{gan2024mgtr,
  title={{MGTR: Multi-Granular Transformer for Motion Prediction with LiDAR}},
  author={Gan, Yiqian and Xiao, Hao and Zhao, Yizhe and Zhang, Ethan and Huang, Zhe and Ye, Xin and Ge, Lingting},
  booktitle=icra,
  year={2024}
}

@inproceedings{cui2023gorela,
  title={{GoRela: Go Relative for Viewpoint-Invariant Motion Forecasting}},
  author={Cui, Alexander and Casas, Sergio and Wong, Kelvin and Suo, Simon and Urtasun, Raquel},
  booktitle=icra,
  IGNOREpages={7801--7807},
  year={2023},
}

@article{zhang2024simpl,
  title={{SIMPL: A Simple and Efficient Multi-agent Motion Prediction Baseline for Autonomous Driving}},
  author={Zhang, Lu and Li, Peiliang and Liu, Sikang and Shen, Shaojie},
  journal={RAL},
  year={2024},
}

@inproceedings{zhang2024demo,
 title={{DeMo: Decoupling Motion Forecasting into Directional Intentions and Dynamic States}},
 author={Zhang, Bozhou and Song, Nan and Zhang, Li},
 booktitle=NIPS,
 year={2024},
}

@inproceedings{song2024realmotion,
 title={{Motion Forecasting in Continuous Driving}},
 author={Song, Nan and Zhang, Bozhou and Zhu, Xiatian and Zhang, Li},
 booktitle=NIPS,
 year={2024},
}

@inproceedings{zeng2021lanercnn,
  title={{LaneRCNN: Distributed Representations for Graph-Centric Motion Forecasting}},
  author={Zeng, Wenyuan and Liang, Ming and Liao, Renjie and Urtasun, Raquel},
  booktitle=iros,
  IGNOREpages={532--539},
  year={2021},
}

@inproceedings{nayakanti2023wayformer,
  title={{Wayformer: Motion Forecasting via Simple \& Efficient Attention Networks}},
  author={Nayakanti, Nigamaa and Al-Rfou, Rami and Zhou, Aurick and Goel, Kratarth and Refaat, Khaled S and Sapp, Benjamin},
  booktitle=icra,
  IGNOREpages={2980--2987},
  year={2023},
}

@inproceedings{qi2017pointnet,
  title={{PointNet: Deep Learning on Point Sets for 3D Classification and Segmentation}},
  author={Qi, Charles R and Su, Hao and Mo, Kaichun and Guibas, Leonidas J},
  booktitle=cvpr,
  IGNOREpages={652--660},
  year={2017}
}

@inproceedings{loshchilov2017decoupled,
  title={{Decoupled Weight Decay Regularization}},
  author={Loshchilov, Ilya and Hutter, Frank},
  booktitle=iclr,
  year={2019}
}

@article{huber1964robust,
  title={{Robust Estimation of a Location Parameter}},
  author={Huber, Peter J},
  journal={{The Annals of Mathematical Statistics}},
  year={1964},
}

@inproceedings{mercat2020multi,
  title={{Multi-Head Attention for Multi-Modal Joint Vehicle Motion Forecasting}},
  author={Mercat, Jean and Gilles, Thomas and El Zoghby, Nicole and Sandou, Guillaume and Beauvois, Dominique and Gil, Guillermo Pita},
  booktitle=icra,
  IGNOREpages={9638--9644},
  year={2020},
}

@inproceedings{carion2020end,
  title={{End-to-End Object Detection with Transformers}},
  author={Carion, Nicolas and Massa, Francisco and Synnaeve, Gabriel and Usunier, Nicolas and Kirillov, Alexander and Zagoruyko, Sergey},
  booktitle=eccv,
  IGNOREpages={213--229},
  year={2020},
}

@inproceedings{wang2023prophnet,
  title={{ProphNet: Efficient Agent-Centric Motion Forecasting with Anchor-Informed Proposals}},
  author={Wang, Xishun and Su, Tong and Da, Fang and Yang, Xiaodong},
  booktitle=cvpr,
  IGNOREpages={21995--22003},
  year={2023}
}

@inproceedings{liang2020learning,
  title={{Learning Lane Graph Representations for Motion Forecasting}},
  author={Liang, Ming and Yang, Bin and Hu, Rui and Chen, Yun and Liao, Renjie and Feng, Song and Urtasun, Raquel},
  booktitle=eccv,
  IGNOREpages={541--556},
  year={2020},
}

@inproceedings{varadarajan2022multipath,
  title={{MultiPath++: Efficient Information Fusion and Trajectory Aggregation for Behavior Prediction}},
  author={Varadarajan, Balakrishnan and Hefny, Ahmed and Srivastava, Avikalp and Refaat, Khaled S and Nayakanti, Nigamaa and Cornman, Andre and Chen, Kan and Douillard, Bertrand and Lam, Chi Pang and Anguelov, Dragomir and others},
  booktitle=icra,
  IGNOREpages={7814--7821},
  year={2022},
}

@inproceedings{chang2019argoverse,
  title={{Argoverse: 3D Tracking and Forecasting with Rich Maps}},
  author={Chang, Ming-Fang and Lambert, John and Sangkloy, Patsorn and Singh, Jagjeet and Bak, Slawomir and Hartnett, Andrew and Wang, De and Carr, Peter and Lucey, Simon and Ramanan, Deva and others},
  booktitle=cvpr,
  IGNOREpages={8748--8757},
  year={2019}
}

@article{liu2021multimodal,
  title={{Multimodal Motion Prediction with Stacked Transformers}},
  author={Liu, Yicheng and Zhang, Jinghuai and Fang, Liangji and Jiang, Qinhong and Zhou, Bolei},
  journal=cvpr,
  IGNOREpages={7577--7586},
  year={2021}
}

@inproceedings{zhang2023hptr,
  title={{Real-Time Motion Prediction via Heterogeneous Polyline Transformer with Relative Pose Encoding}},
  booktitle=NIPS,
  author={Zhang, Zhejun and Liniger, Alexander and Sakaridis, Christos and Yu, Fisher and Van Gool, Luc},
  year={2023},
}

@inproceedings{shi2022motion,
  title={{Motion Transformer with Global Intention Localization and Local Movement Refinement}},
  author={Shi, Shaoshuai and Jiang, Li and Dai, Dengxin and Schiele, Bernt},
  booktitle=NIPS,
  IGNOREpages={6531--6543},
  year={2022},
}

@inproceedings{lan2023sept,
  title={{SEPT: Towards Efficient Scene Representation Learning for Motion Prediction}},
  author={Lan, Zhiqian and Jiang, Yuxuan and Mu, Yao and Chen, Chen and Li, Shengbo Eben and Zhao, Hang and Li, Keqiang},
  booktitle=iclr,
  year={2023}
}

@inproceedings{cheng2023forecast,
  title={{Forecast-MAE: Self-supervised Pre-training for Motion Forecasting with Masked Autoencoders}},
  author={Cheng, Jie and Mei, Xiaodong and Liu, Ming},
  booktitle=iccv,
  year={2023}
}

@inproceedings {wilson2021argoverse,
  author = {Benjamin Wilson and William Qi and Tanmay Agarwal and John Lambert and Jagjeet Singh and Siddhesh Khandelwal and Bowen Pan and Ratnesh Kumar and Andrew Hartnett and Jhony Kaesemodel Pontes and Deva Ramanan and Peter Carr and James Hays},
  title = {{Argoverse 2: Next Generation Datasets for Self-driving Perception and Forecasting}},
  booktitle = {NeurIPS Datasets and Benchmarks},
  year = {2021}
}

@inproceedings{ettinger2021large,
  title={{Large Scale Interactive Motion Forecasting for Autonomous Driving: The Waymo Open Motion Dataset}},
  author={Ettinger, Scott and Cheng, Shuyang and Caine, Benjamin and Liu, Chenxi and Zhao, Hang and Pradhan, Sabeek and Chai, Yuning and Sapp, Ben and Qi, Charles R and Zhou, Yin and others},
  booktitle=iccv,
  IGNOREpages={9710--9719},
  year={2021}
}

@inproceedings{caesar2020nuscenes,
  title={{nuScenes: A Multimodal Dataset for Autonomous Driving}},
  author={Caesar, Holger and Bankiti, Varun and Lang, Alex H and Vora, Sourabh and Liong, Venice Erin and Xu, Qiang and Krishnan, Anush and Pan, Yu and Baldan, Giancarlo and Beijbom, Oscar},
  booktitle=cvpr,
  IGNOREpages={11621--11631},
  year={2020}
}

@inproceedings{huang2025trajectory,
  title={{Trajectory Mamba: Efficient Attention-Mamba Forecasting Model Based on Selective SSM}},
  author={Huang, Yizhou and Cheng, Yihua and Wang, Kezhi},
  booktitle=cvpr,
  year={2025}
}

@inproceedings{demmler2025dynamic,
  title={{Dynamic Intent Queries for Motion Transformer-based Trajectory Prediction}},
  author={Demmler, Tobias and Hartung, Lennart and Tamke, Andreas and Dang, Thao and Hegai, Alexander and Haug, Karsten and Mikelsons, Lars},
  booktitle=iv,
  year={2025}
}
}

\clearpage \appendix 
\maketitlesupplementary
In this supplementary, we first present a comparison of our streaming processing to traditional snapshot-based methods in \Cref{sec:app:stream}.
To support reproducibility, in the following, we provide all implementation details of our model (\Cref{sec:app:arch_details}) and how to apply it to the multi-agent setting (\Cref{sec:app:multiagent}).
We further detail the evaluations (\Cref{sec:app:evaldetails}), present additional results, such as a robustness study on recovering early prediction errors, and provide visualizations of our approach (\Cref{sec:app:results}).
Finally, we provide an overview of the code framework (\Cref{sec:app:code}) which is also included as supplemental material.

\section{Streaming vs. Snapshot-based Processing}
\label{sec:app:stream}
\begin{figure}[tp]
    \centering
    \includegraphics[trim={0cm, 0cm, 0cm, 0cm}, clip, width=\linewidth]{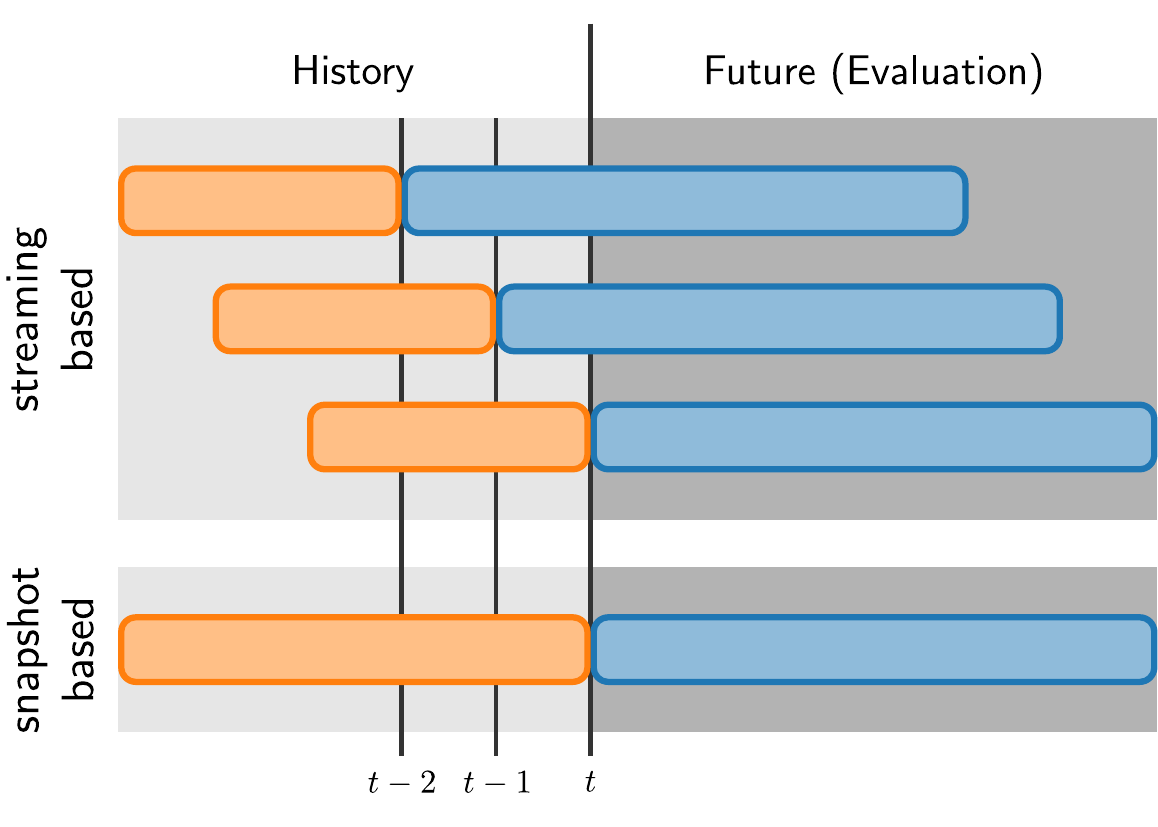}
    \vspace{-0.6cm}
    \caption{
    Comparison between snapshot-based and streaming-based trajectory prediction paradigms.
    Both approaches operate on the same benchmark input data, ensuring a fair comparison without additional data.
    In the streaming paradigm, past observations are processed using a sliding window, closely resembling practical deployment conditions.
    To model the challenges of real-world operation information relay mechanisms are established in streaming processing.
    This enables more consistent and temporally coherent predictions compared to snapshot-based processing, which handles each frame independently.
    In this example, predictions from the third pass of the streaming model can be directly evaluated against those from the snapshot-based approach.
    }
    \vspace{-0.5cm}
    \label{fig:s_stream}
\end{figure}

\Cref{fig:s_stream} compares the standard snapshot-based trajectory prediction paradigm to the streaming-based processing employed by us and related work~\cite{song2024realmotion, zhang2024demo}.
Both paradigms use the same scenario splits to separate training and validation data, ensuring a fair comparison.
To enable streaming processing without longer context data, the historical input fed into the model is reduced.
This allows multiple predictions to be executed within the same set of training data.

The streaming-based processing closely reflects real-world deployment scenarios, where trajectory prediction models operate in a continuous setting.
It enables an information flow between successive prediction steps, unlike the snapshot-based paradigm, where each prediction is performed independently.
This allows models to leverage temporal information, leading to more robust and consistent forecasting.

For Argoverse~2~\cite{wilson2021argoverse} we set the model input history to $h=3\,\mathrm{s}$, which corresponds to $T_h=30$ input samples and we execute three predictions with a $1\,\mathrm{s}$ time gap in between at $t \in \{3, 4, 5\}\,\mathrm{s}$.
Following the benchmark settings, we predict a future of $6\,\mathrm{s}$ which corresponds to $T_f=60$ output time steps.
To fully leverage the available training data during streaming processing, we also predict for \mbox{$T_a=20$} additional steps into the future.
This leads to a total model output of $8\,\mathrm{s}$ seconds.
For all evaluations and also for selecting our endpoint-centric features only the future up to $T_f$ steps is relevant.
During the first and second model pass, the additional future steps $T_a$ allow us to compute the regression losses on a longer future horizon, effectively teaching the model to predict long-term future.
For our ablation on Argoverse~1~\cite{chang2019argoverse} we use a historic input of $1\,\mathrm{s}$ ($T_h=10$) and execute three passes at $t \in \{1, 1.5, 2\}\,\mathrm{s}$.

\section{Implementation Details}
\label{sec:app:arch_details}
We set the feature dimension of our model to $D=128$.
Our categorical type embeddings distinguish between four agent types (vehicles, pedestrians, cyclists, and others) and three lane types (standard lanes, bike lanes and bus lanes).
We use the standard radius of 150 meters to collect scene elements~\cite{cheng2023forecast, prutsch24efficient, zhang2024demo, song2024realmotion}.
To maintain a manageable number of scene tokens, we do not split the lanes into smaller segments as \eg suggested by~\cite{lan2023sept, gan2024mgtr}.
We use the $xy$-positions, $xy$-velocities and a corresponding \emph{valid} flag to encode agent data, leading to $D_a=5$.
To encode the lane data, we use the $xy$-positions of each lane point and a \emph{valid} flag to mask parts of lane segments exceeding the region of interest, leading to $D_l=3$.

\subsection{Positional Embeddings}
To model the global poses of all elements --- agents, lanes, and target-centric coordinate systems --- we represent each using its position $(x, y)$ and rotation $yaw$. For agents, we use their initial position and orientation; for lanes, we use the center point and orientation of the centerline; and for target-centric coordinate systems, we use the trajectory endpoint and compute the orientation based on the curvature at the trajectory's end.

\subsection{Target-Centric Coordinate Systems}
\begin{figure}[tp]
    \centering
    \vspace{0.2cm}
    \includegraphics[trim={0cm, 0cm, 0cm, 0cm}, clip, width=0.5\linewidth]{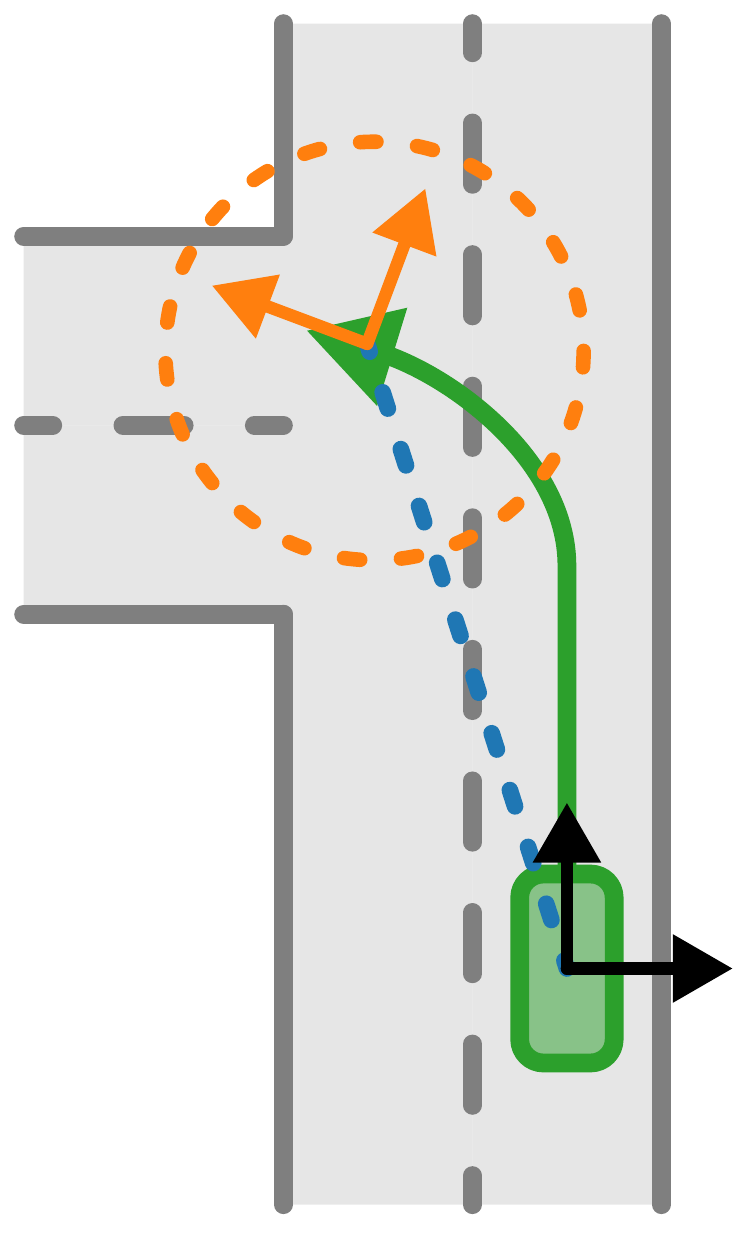}
    \definecolor{darkgreen}{rgb}{0.0, 0.39, 0}
    \caption{
    Modeling of target-centric coordinate system (\textcolor{orange}{orange}) using the prediction (\textcolor{darkgreen}{green}) from the previous time frame.
    We also show the target context radius (\textcolor{orange}{orange}) and the positional embedding (\textcolor{blue}{blue}) used to model the relationship to the agent-centric origin (black).
    }
    \label{fig:s_coord}
\end{figure}

\Cref{fig:s_coord} illustrates the setup of our target-centric coordinate systems.
We use the endpoint from trajectory prediction $F^{t-1}$ at the previous time step to create a local reference frame centered and oriented around the target region.
All scene features within the target context radius are then encoded \wrt the target coordinate system.
To capture global spatial relationships, we additionally incorporate positional embeddings from the agent-centric origin to the target-centric frame (blue dashed line).

\subsection{Multi-Head Attention Blocks}
For our multi-head attention blocks we use the following configuration:
\begin{itemize}
    \item Number of attention heads: 8
    \item Dropout rate: 0.2
    \item Feedforward expansion factor: $4\cdot D$
    \item Activation function: GELU~\cite{hendrycks2016gaussian}
    \item Norm before attention operation
\end{itemize}

\subsection{Encoder}
We utilize four encoder blocks in the agent encoder $f_A$ and four encoding blocks in our scene context encoder $f_S$.
For our target context encoder $f_T$ we utilize two encoder blocks.
\Cref{tab:abl_enc_depth} provides an ablation study for the depth of our novel target context encoder.
Since the number of tokens $N'_c$ in our target-centric context is smaller than the number of tokens $N_c$ in the agent-centric context, a shallower encoder can be used.
To compute the position embeddings for a given pose $(x, y, \theta)$ we first transform it to $(x, y, \sin \theta, \cos \theta)$.
Then, we apply a two-layer multilayer perceptron~(MLP):
\begin{itemize}
    \item Layer 1: $4 \times D$
    \item GELU~\cite{hendrycks2016gaussian} activation function
    \item Layer 2: $D \times D$
\end{itemize}

\subsection{Decoder}
Our dual-context decoder consists of three stages ($d=3$), resulting in a total of six cross-attention blocks.
If an agent has no previous prediction, the cross-attention step to the target-centric context is skipped.
The MLP to process the decoded queries $Q'$ and output future trajectories $F$ has two layers:
\begin{itemize}
    \item Layer 1: $D \times 2\cdot  D$
    \item ReLU activation function
    \item Layer 2: $2\cdot D \times 2\cdot (T_f + T_a)$
\end{itemize}
Similarly, the MLP for outputting probability scores $P$:
\begin{itemize}
    \item Layer 1: $D \times 2\cdot  D$
    \item ReLU activation function
    \item Layer 2: $2\cdot D \times 1$
\end{itemize}

As described above~(\Cref{{sec:app:stream}}), our decoder predicts $T_a$ additional steps in addition to the $T_f$ steps which are used as actual trajectories in the evaluation.

\subsection{Optimization}
We employ a common winner-takes-all strategy for training our model~\cite{cheng2023forecast}, where only the best fitting predicted trajectory (\ie with the smallest average displacement error) is used for optimization.
We utilize a smooth L1 loss~(Huber loss~\cite{huber1964robust}) as regression loss $L_{\text{reg}}$ to ensure that the hypothesis fits to the ground truth.
Additionally, we use a standard cross-entropy loss $L_{\text{cls}}$ to assign the highest confidence score to the best-fitting trajectory.
To further enhance learning, we employ an auxiliary loss $L_{\text{aux}}$~\cite{cheng2023forecast}, where a single trajectory is predicted for each non-focal agent present in the scene, and a smooth L1 loss is applied.
To predict the auxiliary future, we utilize a linear layer $D\times 2\cdot T_f$ for each token in our \mbox{scene context $S$} that corresponds to an agent excluding the focal agent.
The final loss is given as $L=L_{\text{reg}}+L_{\text{cls}}+L_{\text{aux}}$.

\begin{table}[t]
\centering
\setlength{\tabcolsep}{6pt}
\footnotesize{
\begin{tabular}{cc|cccc}
\toprule
Dataset & \# Blocks & mADE$_6$ & mFDE$_6$ & b-mFDE$_6$ \\ 
\midrule
& 1 & \bval{0.66} & 1.26 & 1.86      \\ %
AV2 Val & \cellcolor{rcol}2 & \cellcolor{rcol}\bval{0.66} & \cellcolor{rcol}\bval{1.25} & \cellcolor{rcol}\bval{1.85}  \\ %
& 3 & \bval{0.66} & 1.26 & \bval{1.85}  \\ %
\bottomrule
\end{tabular}
}
\caption{
Ablation study for different number of attention blocks in our target encoder $f_T$.
}
\vspace{-0.05in}
\label{tab:abl_enc_depth}
\end{table}

We train our model for 80 epochs, with the first 13 epochs serving as warm-up phase, during which the we linearly increase the learning rate from 1e-5 to 1e-2.
Afterward, we decrease the learning rate to 1e-5 using a single cosine annealing schedule.
Training is executed using a batch size of 32 on a single NVIDIA Quadro RTX 8000.
We utilize AdamW~\cite{loshchilov2017decoupled} as optimizer, employ norm-based gradient clipping with maximum value set to 5 and execute weight decay with 1e-2.
We only train on the Argoverse~2 training set without any pre-training or data augmentations.

\section{Multi-Agent Extension}
\label{sec:app:multiagent}
\subsection{Implementation Details}
In the AV2 multi-agent settings the goal is to predict trajectories for all scored agents.
Our global consistency module employs two transformer blocks for self-attention across all modes of an agent, followed by two transformer blocks for self-attention across agents per mode (world).
In the first stage, we do self-attention on our decoded $Q' \in \mathbb{R}^{N_{a,s} \times K \times D}$ across all modes $K$ for each agent, where $N_{a,s}$ is the number of scored agents.
Next, we permute the queries to $\mathbb{R}^{K \times N_{a,s} \times D}$  and perform self-attention across all agents $N_{a,s}$ per mode.
To better model different agent behavior, we add categorical type embeddings to $Q'$, distinguishing between focal-agents, \emph{scored}-agents which are driving and \emph{scored}-agents that are likely to be parked (based on the marginal predictions).
We generate the world predictions  $F_w \in \mathbb{R}^{K \times N_a,s \times T_f \times 2}$ using a two-layer MLP:
\begin{itemize}
    \item Layer 1: $D \times 2\cdot  D$
    \item ReLU activation function
    \item Layer 2: $2\cdot D \times 2\cdot T_f$
\end{itemize}
Similarly, the MLP for outputting the associated confidence scores $P_w \in \mathbb{R}^K$:
\begin{itemize}
    \item Layer 1: $D \times 2\cdot  D$
    \item ReLU activation function
    \item Layer 2: $2\cdot D \times 1$
\end{itemize}
We use the same streaming-processing setup as in the single-agent setting.

\subsection{Optimization}
To adapt our approach to the multi-agent setting in Argoverse~2~(AV2)~\cite{wilson2021argoverse}, we initialize the model weights using our single-agent model (marginal prediction model).
We then jointly train the marginal prediction model and the global consistency module end-to-end on the AV2 multi-agent training set for 35 epochs, without any warm-up phase.
To reduce memory consumption, we freeze the agent and lane encoders.
During training, we apply a cosine annealing schedule to decay the learning rate from 1e-2 to 1e-5.

Again, we employ a winner-takes-all principle by optimizing only the trajectories of the best world (lowest average minimum displacement error).
Following single-agent training, we use a regression loss $L_{\text{reg}}$ for each agent prediction in the best world and employ a cross-entropy classification loss $L_{\text{cls}}$ to assign the highest confidence score to the best-fitting world.
To retain strong single-agent performance and guide multi-agent learning, we also include the single-agent losses $L_\text{marginal}$ (see above) during training on the multi-agent setting.
The final loss is given as $L=L_{\text{reg}}+L_{\text{cls}}+L_\text{marginal}$.

\section{Evaluation Details}
\label{sec:app:evaldetails}

\subsection{Metrics}
We evaluate our approach using the standard AV2 benchmark metrics.
Each metric is evaluated using the top $k$ scoring trajectory hypotheses.
The minimum average displacement error (minADE$_k$) is the mean Euclidean distance between the ground truth and the best-fitting hypotheses across all time steps. The minimum final displacement error (minFDE$_k$) considers only the distance at the final time step, while the Brier minimum final displacement error (brier-minFDE$_k$) adds a penalty term $(1-\pi)^2$ to the minFDE$_k$, where $\pi$ is the probability score for the best-fitting trajectory.
The miss rate (MR$_k$) evaluates whether any predicted endpoint is within a radius of 2 meters from the ground truth endpoint.

In the multi-agent setting, we evaluate all metrics for the top $k$ scoring worlds.
Each world contains one future trajectory for each \emph{scored} agent.
The average minimum average displacement error (avgMinADE$_k$) is computed by averaging the minADE$_k$ across all actors within a world.
Analogously, the average minimum final displacement error (avgMinFDE$_k$) considers the minFDE$_k$ for all actors, and the average brier minimum final displacement error (avgBrierMinFDE$_k$) averages the brier-minFDE$_k$ across all actors in a world.
The actor miss rate actorMR$_k$ denotes the rate of all scored agents which have an endpoint within 2 meters around the ground truth endpoint.

\subsection{Latency Measurements}
We utilize the official code implementations to measure the latencies of QCNet\footnote{\url{https://github.com/ZikangZhou/QCNet}}~\cite{zhou2023query},
RealMotion\footnote{\url{https://github.com/fudan-zvg/RealMotion}\label{fn:remo}}~\cite{song2024realmotion}, and
DeMo\footnote{\url{https://github.com/fudan-zvg/DeMo}}~\cite{zhang2024demo}.
To ensure fair comparison and eliminate influences by different data loading and preprocessing implementations, we measure only the inference latency of the model forward pass.

\begin{table*}[t]
\footnotesize
\centering
\begin{tabular}{l|ccccccc}
\toprule
Method & minADE$_1$ & minFDE$_1$ & MR$_6$ & minADE$_6$ & minFDE$_6$ & brier-minFDE$_\mathbf{6}$ \\ 
\midrule
RealMotion~\cite{song2024realmotion}            & 1.65 & 4.10 & 0.16 & 0.67 & 1.30 & 1.94 \\
DeMo~\cite{zhang2024demo}                       & \bval{1.48} & \bval{3.73} & \bval{0.13} & \bval{0.61} & \bval{1.19} & \sval{1.86} \\
    \rowcolor{rcol}\mn~(Ours)                       & \sval{1.60} & \sval{3.96} & \sval{0.15} & \sval{0.66} & \sval{1.25} & \bval{1.85} \\
\bottomrule
\end{tabular}
\caption{
Comparison of streaming-based methods for single-agent trajectory prediction on the Argoverse~2 validation set.
For all models we report results without model ensembling.
}
\vspace{-0.05in}
\label{tab:av2_val}
\end{table*}

\begin{table}[t]
\centering
\setlength{\tabcolsep}{6pt}
\footnotesize{
\begin{tabular}{c|ccccccc}
\toprule
Endpoint Noise & mADE$_6$ & mFDE$_6$ & b-mFDE$_6$ \\ 
\midrule
$\mathcal{N}(0, 5)$\phantom{-0}  & 0.67 & 1.28 & 1.88 \\  
$\mathcal{U}(-5, 5)$ & 0.67 & 1.27 & 1.87 \\ %
$\mathcal{N}(0, 3)$\phantom{-0} & 0.67 & 1.26 & 1.87 \\  
$\mathcal{U}(-3, 3)$ & 0.67 & 1.26 & 1.86  \\ %
$\mathcal{N}(0, 1)$\phantom{-0} & 0.67 & \bval{1.25} & \bval{1.85}  \\    
$\mathcal{U}(-1, 1)$ & 0.67 & \bval{1.25} & \bval{1.85}  \\   
\cellcolor{rcol}None & \cellcolor{rcol}\bval{0.66} & \cellcolor{rcol}\bval{1.25} & \cellcolor{rcol}\bval{1.85}  \\ %
\bottomrule
\end{tabular}
}
\vspace{-0.1cm}
\caption{
Robustness of our endpoint-aware modeling on the AV2~\cite{wilson2021argoverse} validation set by perturbing the prediction endpoints at $t \in \{3, 4\}$\,s.
These endpoints, which define the anchors for extracting our target-centric features, are modified using additive uniform noise ($\mathcal{U}$) or Gaussian noise ($\mathcal{N}$).
The shown results correspond to the prediction errors at $t=5$s.
}
\vspace{-0.5cm}
\label{tab:errorprop}
\end{table}

\begin{figure}[b]
    \centering
    \includegraphics[trim={0.3cm, 0.1cm, 1.2cm, 1.5cm}, clip, width=0.8\linewidth]{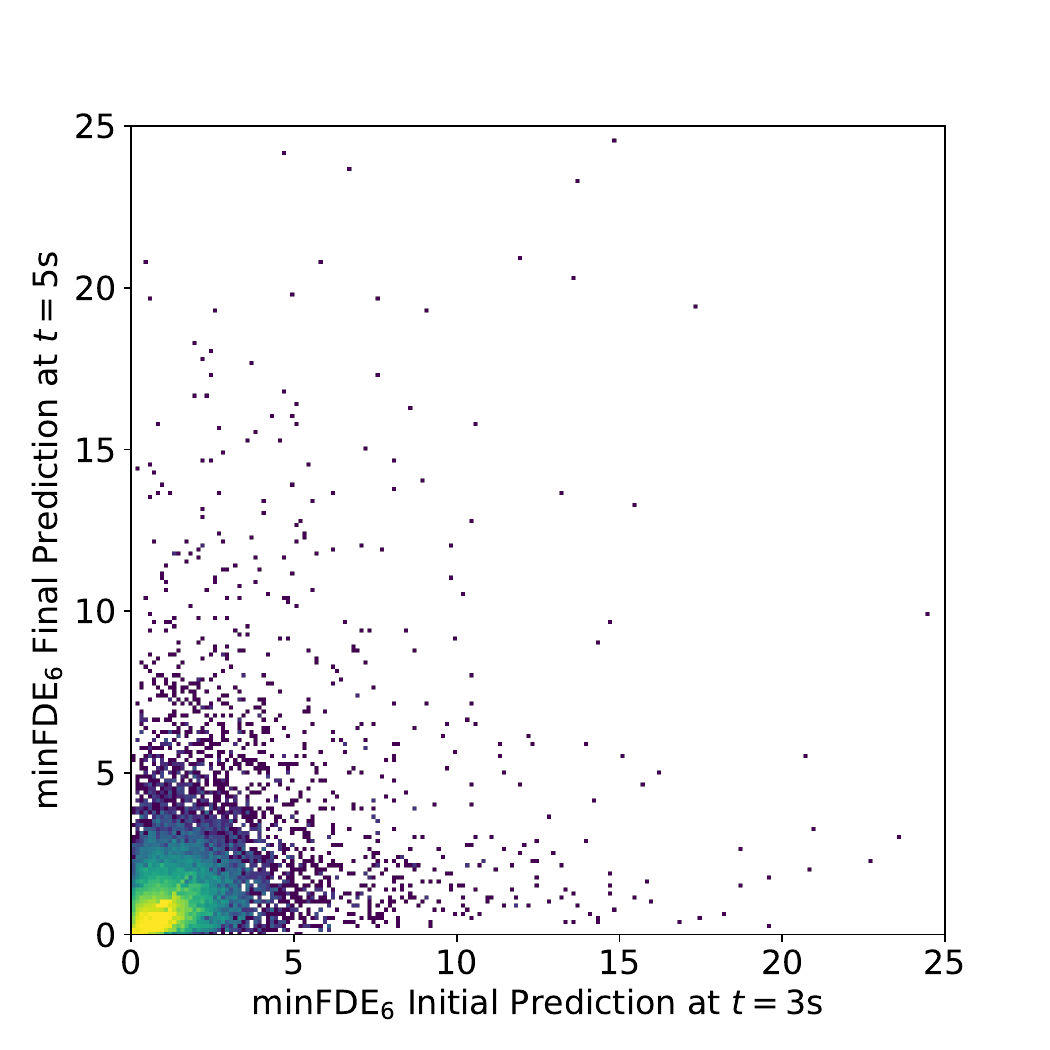} 
    \includegraphics[trim={0.3cm, 0.1cm, 1.2cm, 1.5cm}, clip, width=0.8\linewidth]{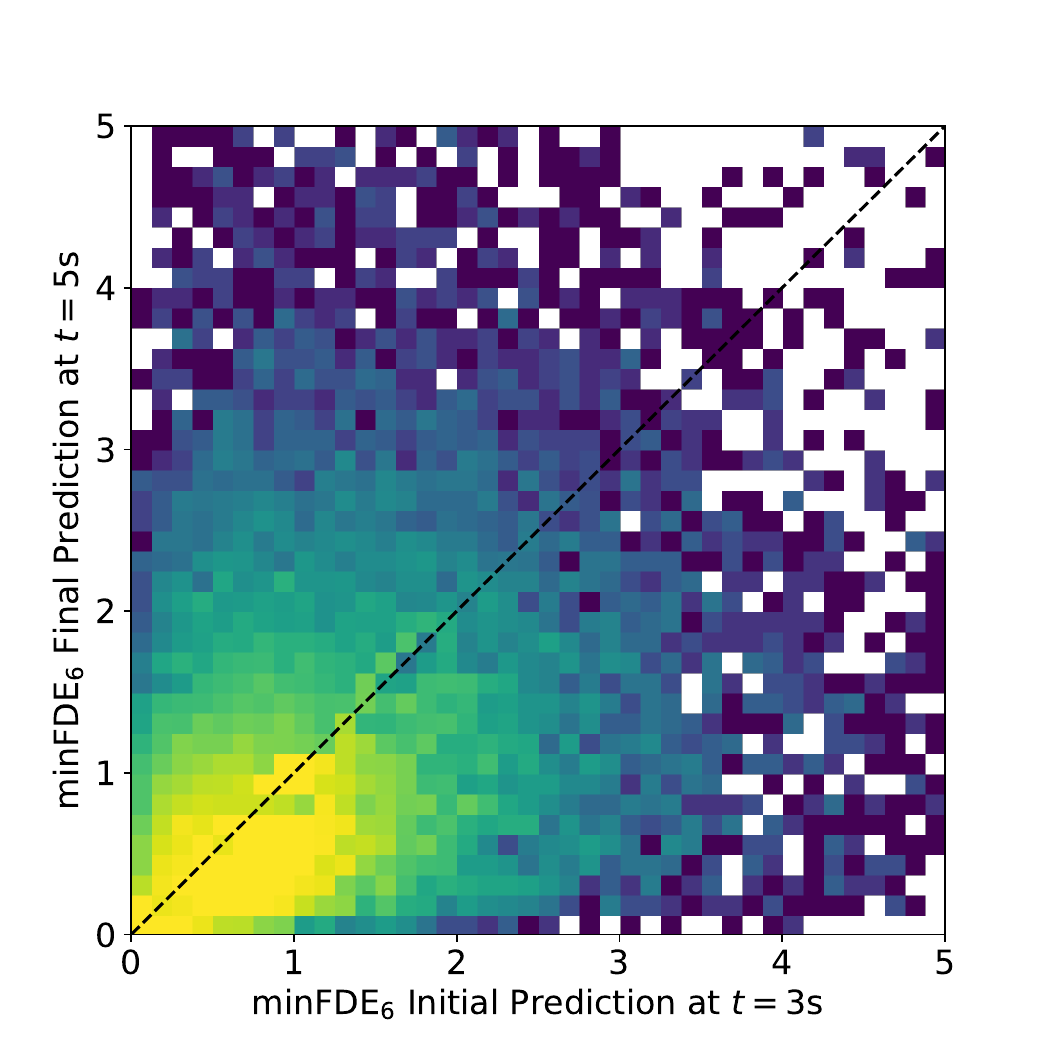}
    \vspace{-0.3cm}
    \caption{
    Density plots of prediction errors.
    Both plots compare the first prediction done at $t=3$\,s (x-axis) versus the final prediction done at at $t=5$\,s (y-axis).
    The top plot shows that even when the initial error is large, the model can recover and correct its prediction.
    The bottom plot provides a zoomed-in view near the origin.
    The higher density of points below the $x = y$ diagonal (dashed line) indicates that predictions generally improve over time.
    }
    \label{fig:s_error}
\end{figure}

\section{Additional Results}
\label{sec:app:results}
\subsection{Streaming Methods on AV2 Validation Set}
\Cref{tab:av2_val} compares streaming-based trajectory prediction approaches on the Argoverse~2 validation set.
Our approach again yields the best brier-minFDE$_6$, which considers both displacement errors and trajectory scoring.
By comparing the brier-minFDE$_6$ to minFDE$_6$ values of the different models, we can assess that our approach excels in estimating the likelihood of predicted trajectories, which is important for interpreting the results in downstream tasks like ego-motion planning.

\subsection{Robustness to Error Propagation}

Table~\ref{tab:errorprop} evaluates the robustness of our endpoint-aware modeling for information streaming when previous predictions are noisy.
To this end, we perturb the endpoint anchors, which are used to obtain our target-centric features, with uniform or Gaussian noise.
The results show that performance only slightly deteriorates under small prediction noise. 
Nevertheless, our dual-context approach maintains robust performance even when past predictions are affected by errors.
The target-centric features provide additional information that improves overall prediction accuracy without constraining predictions, allowing recovery when agent-centric features at the current timestep indicate a different future movement.
As the noise magnitude increases, performance declines further; however, the model remains resilient even under substantial noise offsets.

Furthermore, we analyze the evolution of errors in Figure~\ref{fig:s_error}.
The plots compare the minFDE$_6$ for the first prediction, made at $t=3$\,s into the scenario, with the final prediction at $t=5$\,s, each evaluated over a future horizon of $6$\,s.
The results show that our dual-context decoding approach can recover from early prediction errors: even when the initial prediction is inaccurate (high value on the x-axis), the final prediction can still be highly accurate (low value on the y-axis).
There are also some cases where the initial prediction is accurate but the final prediction degrades, for example when an unexpected maneuver, \eg a sudden stop, is not apparent in the first prediction window but becomes relevant in the second.
Overall, the model achieves improved prediction quality over time, as reflected by the higher point density below the $x=y$ diagonal.

\subsection{Latency on NVIDIA A10 GPU}
We provide additional latency results using one NVIDIA A10 GPU, comparing streaming-based methods in \Cref{tab:latencya10}.
Also on this newer GPU architecture our approach achieves the best latency results, obtaining the lowest online and offline latency across all batch sizes.

\begin{table*}[t]
\centering
\footnotesize{
\begin{tabular}{l|cc|cc|cc}
\toprule
Method & \multicolumn{2}{c|}{Latency ($B=1$)} & \multicolumn{2}{c|}{Latency ($B=32$)} & \multicolumn{2}{c}{Latency ($B=64$)}  \\ 
       & Offline & Online & Offline & Online & Offline & Online \\
\midrule
RealMotion~\cite{song2024realmotion}     & \underline{51~ms} & \textbf{13~ms} & 255~ms & 85~ms & 512~ms & 172~ms     \\
DeMo~\cite{zhang2024demo}                & 73~ms & \underline{18~ms} & \underline{223~ms} & \underline{71~ms} &  \underline{433~ms} & \underline{143~ms}\\
\rowcolor{rcol}\mn~(Ours)                & \textbf{50~ms} & \textbf{13~ms} & \textbf{95~ms} & \textbf{39~ms} & \textbf{185~ms} & \phantom{0}\textbf{65~ms} \\
\bottomrule
\end{tabular}
}
\caption{
Latency analysis for predicting $B$ Argoverse~2 single-agent scenarios using one NVIDIA A10 GPU.
We compare streaming-based methods and report the offline and online inference latency. The online latency is relevant for practical application.
}
\vspace{-0.05in}
\label{tab:latencya10}
\end{table*}

\begin{table*}[t]
\centering
\setlength{\tabcolsep}{3pt}
\footnotesize{
\begin{tabular}{ccc|cccc}
\toprule
Model & Training Data & Global Consistency Module & actorMR$_6$ & avgMinADE$_6$ & avgMinFDE$_6$ & avgBrierMinFDE$_6$ \\ 
\midrule
RealMotion~\cite{song2024realmotion} & Single-Agent Data & \xmark & 0.727 & 1.413 & 4.060 & 4.716 \\ %
\midrule
SEAM (Ours) & Single-Agent Data & \xmark                          & 0.693 & 1.324 & 3.643 & 4.256 \\ %
SEAM (Ours)  & Finetuned on Multi-Agent Data & \xmark             & 0.228 & 0.720 & 1.619 & 2.126 \\ %
\rowcolor{rcol} SEAM (Ours) & Finetuned on Multi-Agent Data & \cmark & \bval{0.155} & \bval{0.600} & \bval{1.180} & \bval{1.814} \\ %
\bottomrule
\end{tabular}
}
\vspace{-0.1cm}
\caption{
Ablation study on adapting single-agent methods to the AV2 multi-agent validation set.
The first two rows show that, without any model adaptations or finetuning, both our approach and related work~\cite{song2024realmotion} perform poorly in the multi-agent setting.
Finetuning on the multi-agent dataset significantly improves performance, and incorporating a global consistency module for scene-level scoring yields the best results.
}
\vspace{-0.05in}
\label{tab:abl_ma}
\end{table*}

\begin{table}[t]
\centering
\footnotesize{
\begin{tabular}{c|ccccccc}
\toprule
Method                               & Fluctuation \\ 
\midrule
RealMotion~\cite{song2024realmotion} & 0.347 \\
\rowcolor{rcol} \mn~(Ours)                           & \bval{0.341} \\
\bottomrule
\end{tabular}
}
\caption{
We compare the trajectory fluctuation of our model to RealMotion~\cite{song2024realmotion}. The fluctuation metric defined by~\cite{pang2023streaming} gives an indication on the consistency of trajectories across multiple timesteps.
}
\vspace{-0.05in}
\label{tab:fluctuation}
\end{table}

\subsection{Multi-Agent Ablation Study}
We present an ablation study on extending single-agent approaches to the multi-agent setting in \Cref{tab:abl_ma}.
The first two rows show that naively applying a single-agent model to the multi-agent benchmark yields poor performance.
This is primarily because the multi-agent dataset contains motion patterns that are not well-represented in the single-agent data.
Moreover, these approaches ignore global consistency across agent predictions.
Without a dedicated global consistency module, we generate multiple plausible future worlds by combining the most likely trajectories of each agent into one world, the second most likely into another, and so on.
Finetuning the model on multi-agent data helps capture a broader range of behaviors (row 3), and incorporating an explicit global consistency module further improves performance, achieving the best results (row 4).

\subsection{Trajectory Fluctuation}
\Cref{tab:fluctuation} compares the trajectory fluctuation of our approach with that of RealMotion~\cite{song2024realmotion}.
The fluctuation metric, as defined by~\cite{pang2023streaming} measures the consistency of trajectories across multiple prediction frames.
Our approach achieves a lower fluctuation score than RealMotion, indicating more consistent predictions.

\subsection{Result Visualizations}
\begin{figure*}[tp]
    \centering
    \resizebox{\linewidth}{!}{
        \begin{tikzpicture}[node distance=0mm, inner sep=0mm]
            \addresrow{0}{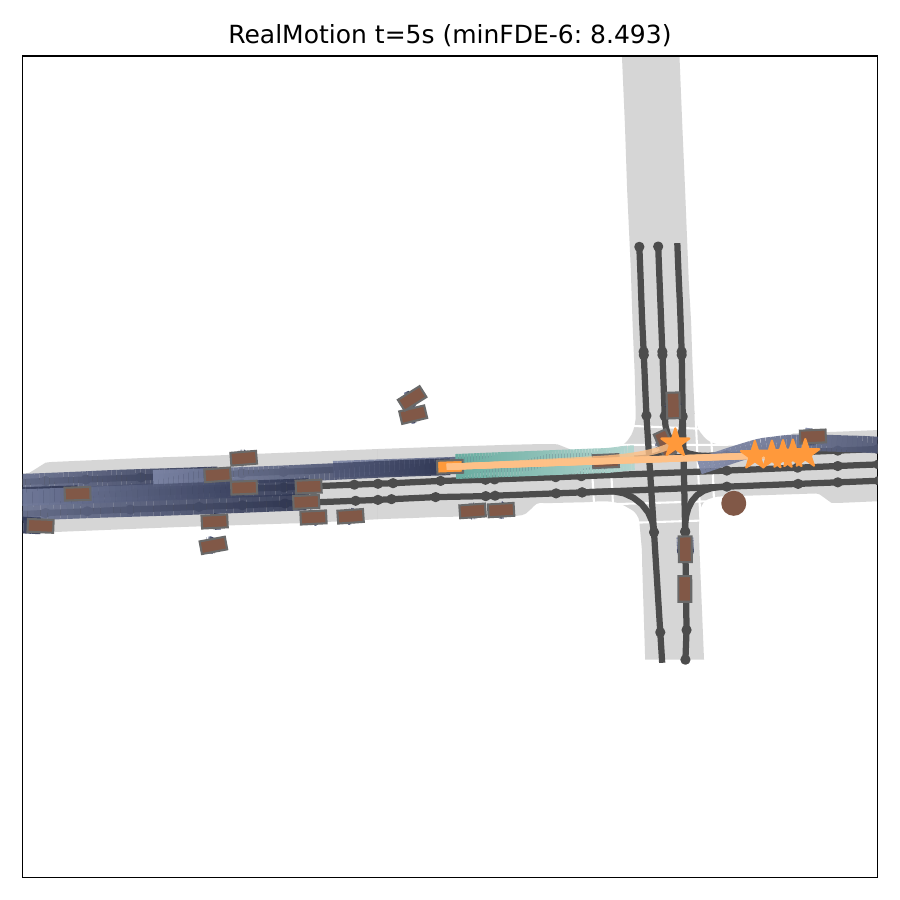}{6}{5}{0}{6}{0.025}{0.01}{}
            \addresrow{1}{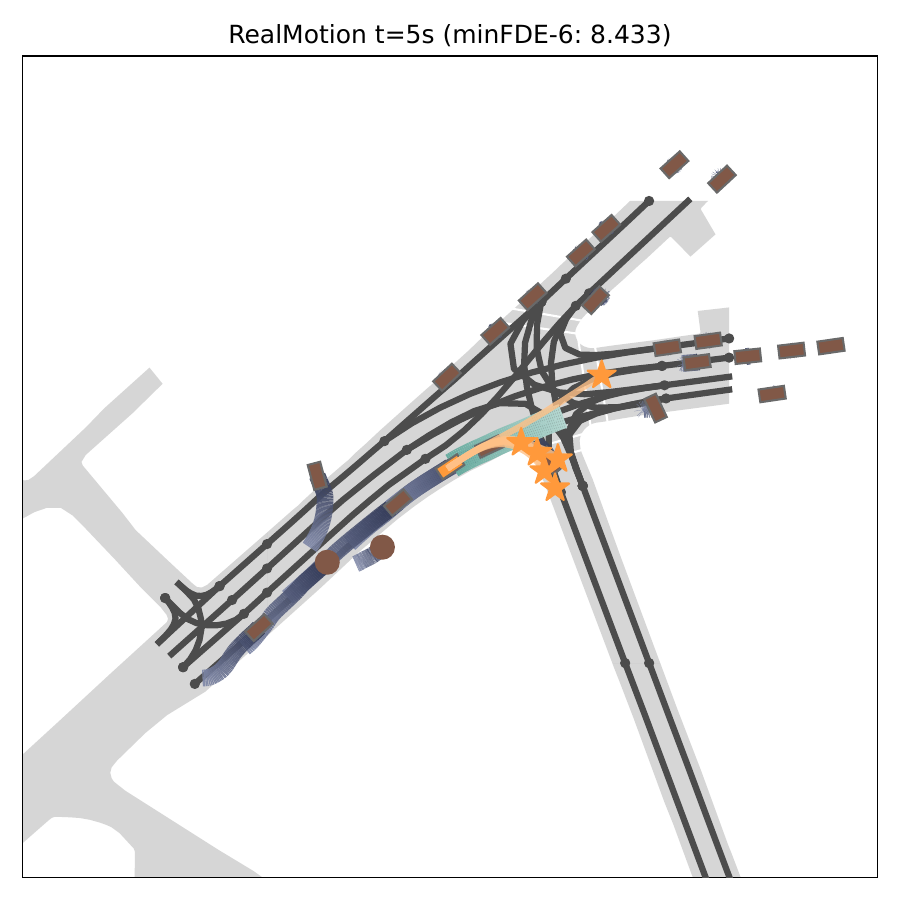}{3}{5.5}{4}{5}{0.025}{0.01}{0}
            \addresrow{2}{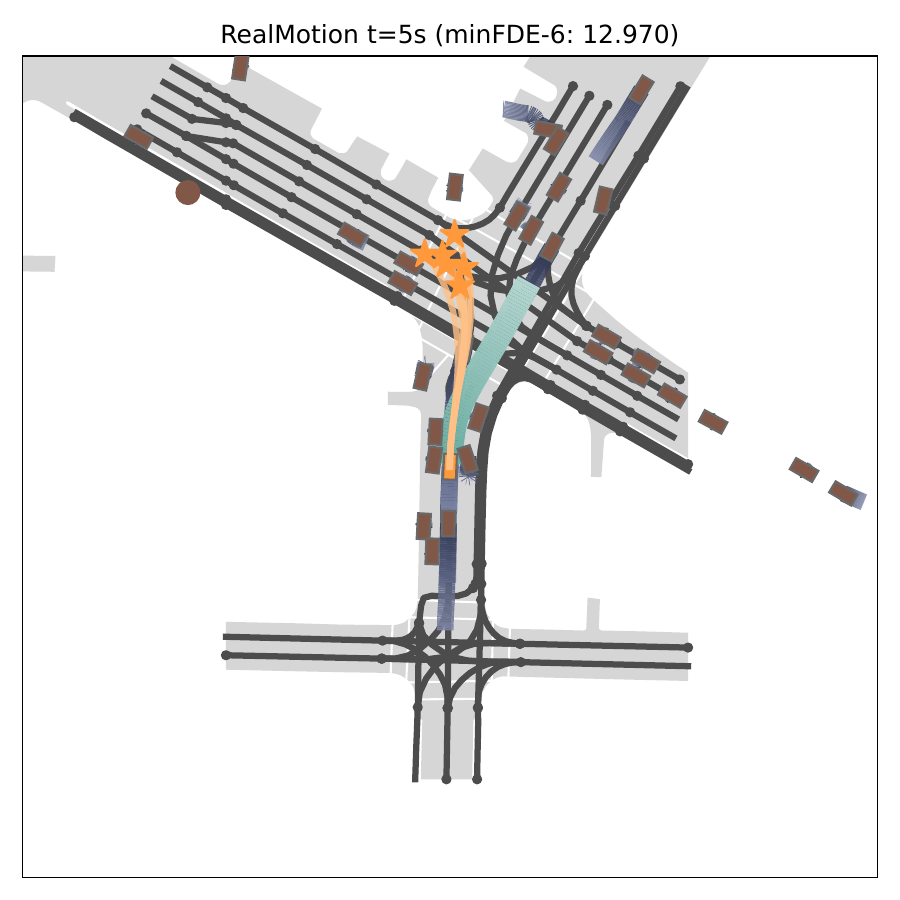}{3}{5.5}{3.5}{2}{0.025}{0.01}{1}
            \addresrow{3}{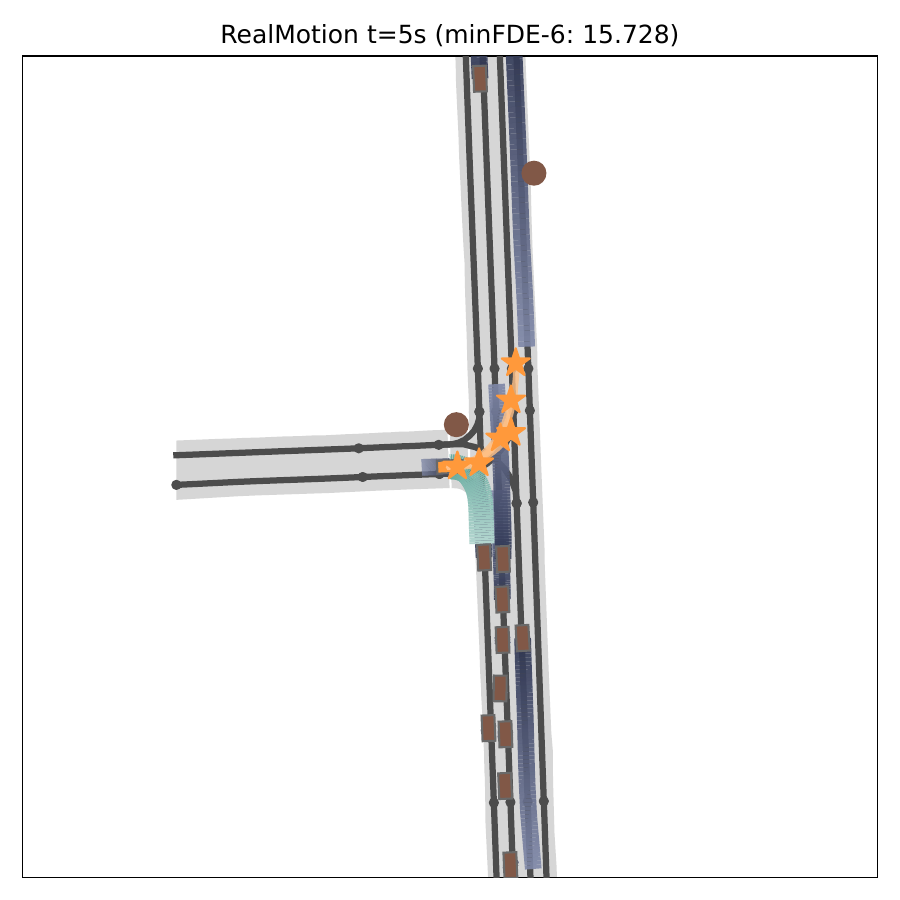}{3}{4.5}{4}{5}{0.025}{0.01}{2}
            \addresrow{4}{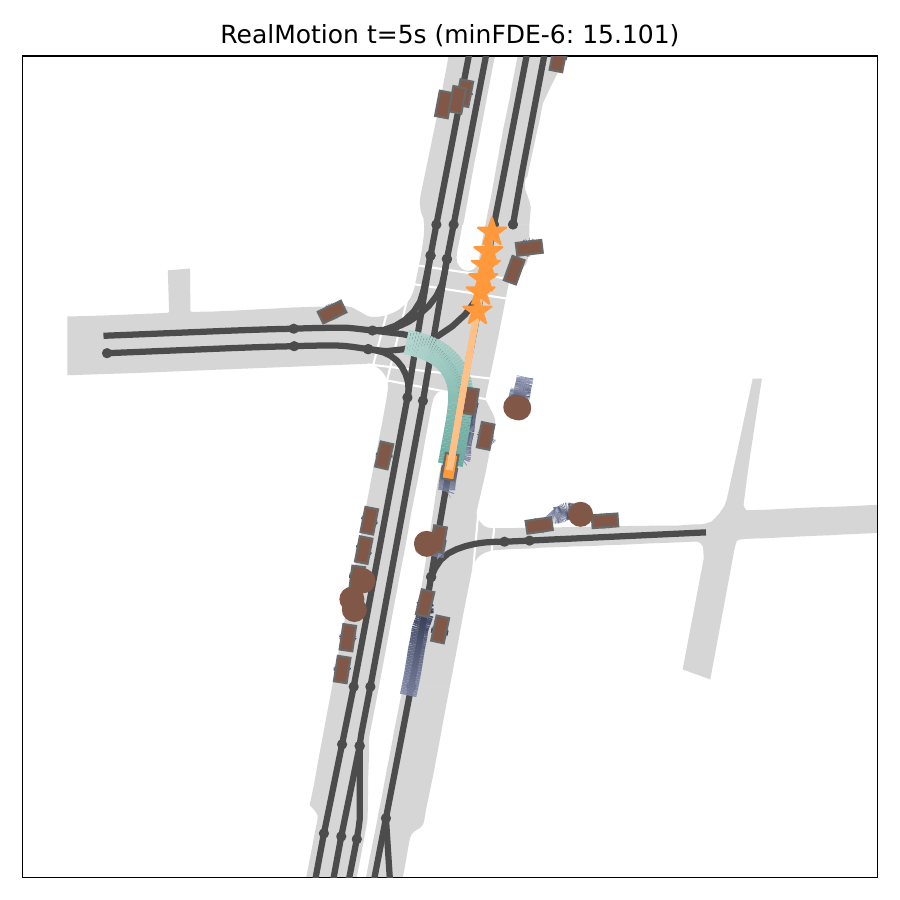}{4}{6}{4}{2.5}{0.025}{0.01}{3}
            \addresrow{5}{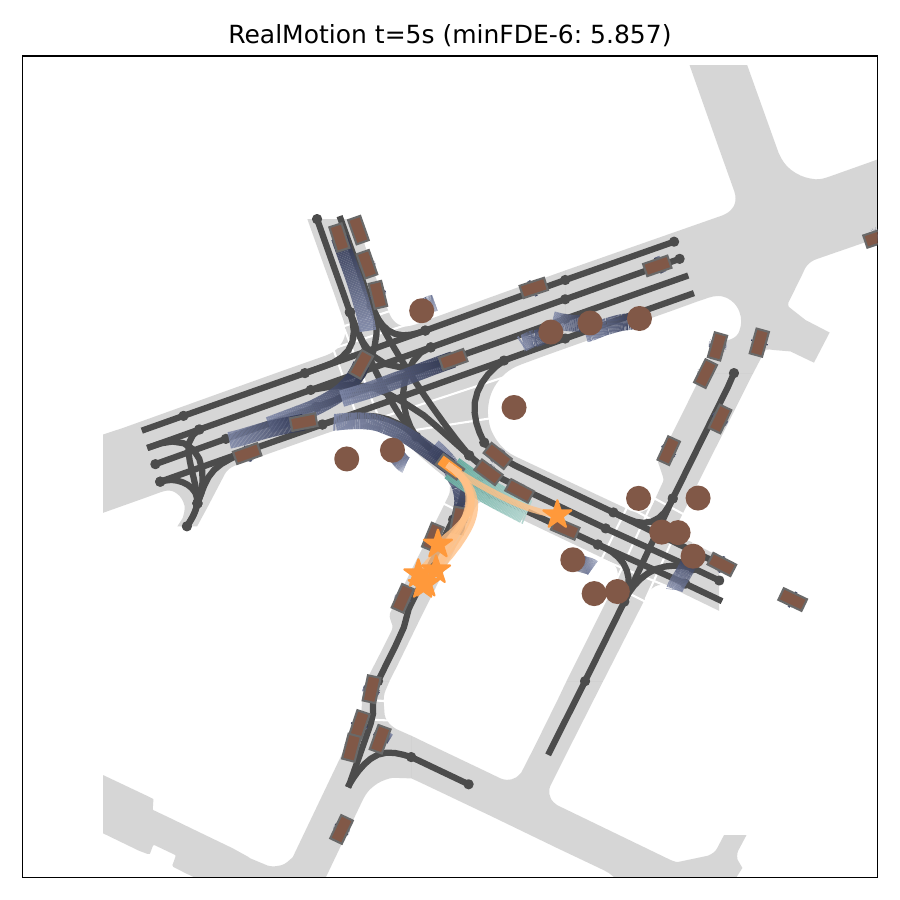}{3}{4}{4}{5.5}{0.025}{0.01}{4}
        \end{tikzpicture}
    }
    \caption{Qualitative results of our approach on scenarios from the Argoverse~2 validation set. We show the \textcolor[HTML]{ff9a3a}{\textbf{predictions}} of our streaming-based method at $t \in \{3, 4, 5\}$s.
    The visualizations also show \textcolor[rgb]{0.41, 0.67, 0.63}{\textbf{ground truth future}}, \textcolor[HTML]{384062}{\textbf{agent histories}}, and \textcolor[HTML]{815847}{\textbf{neighboring agents}}.
    The right column shows the final predictions at $t = 5$s for using RealMotion~\cite{song2024realmotion} in the streaming setting.}
\label{fig:s_results1}
\end{figure*}

\begin{figure*}[tp]
    \centering
    \resizebox{\linewidth}{!}{
    \begin{tikzpicture}[node distance=0mm, inner sep=0mm]
        \addresrow{0}{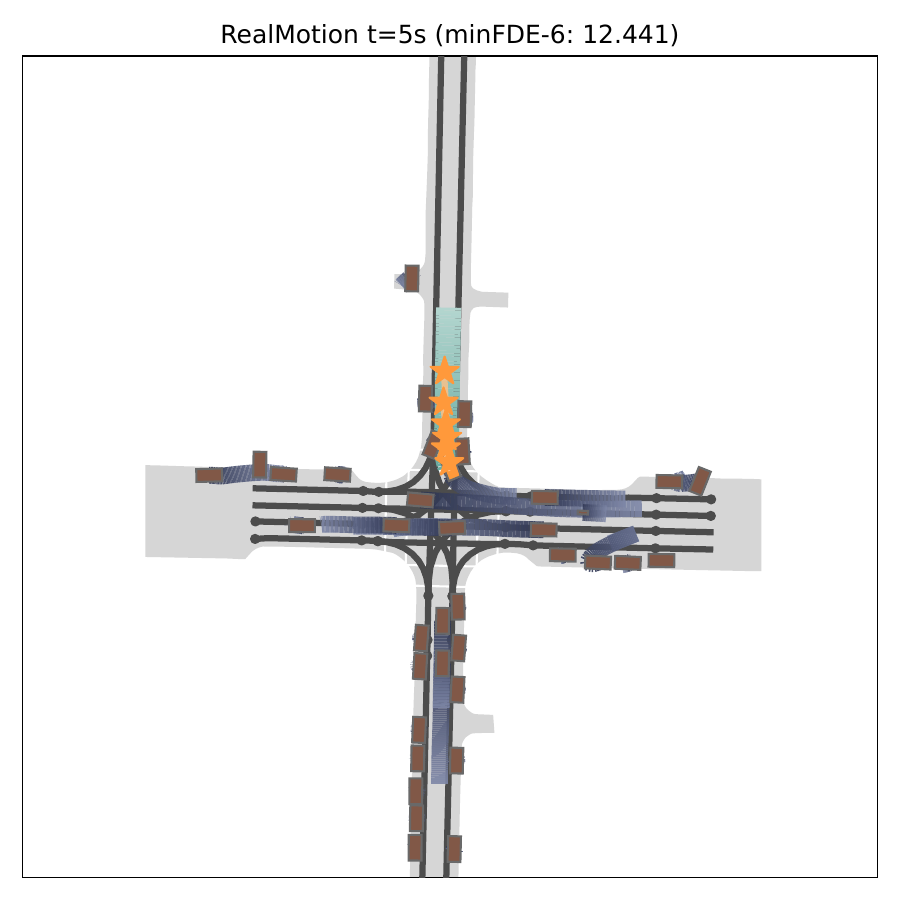}{3}{5}{4}{2.5}{0.025}{0.01}{}
        \addresrow{1}{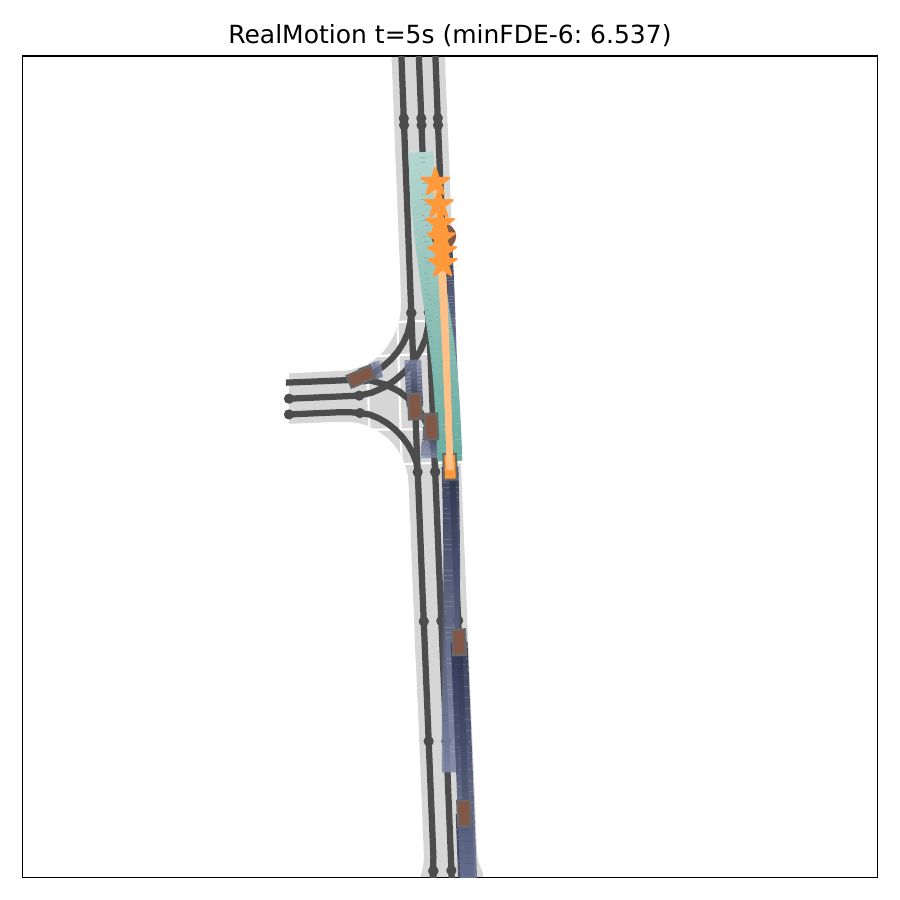}{3}{6}{3}{1}{0.025}{0.01}{0}
        \addresrow{2}{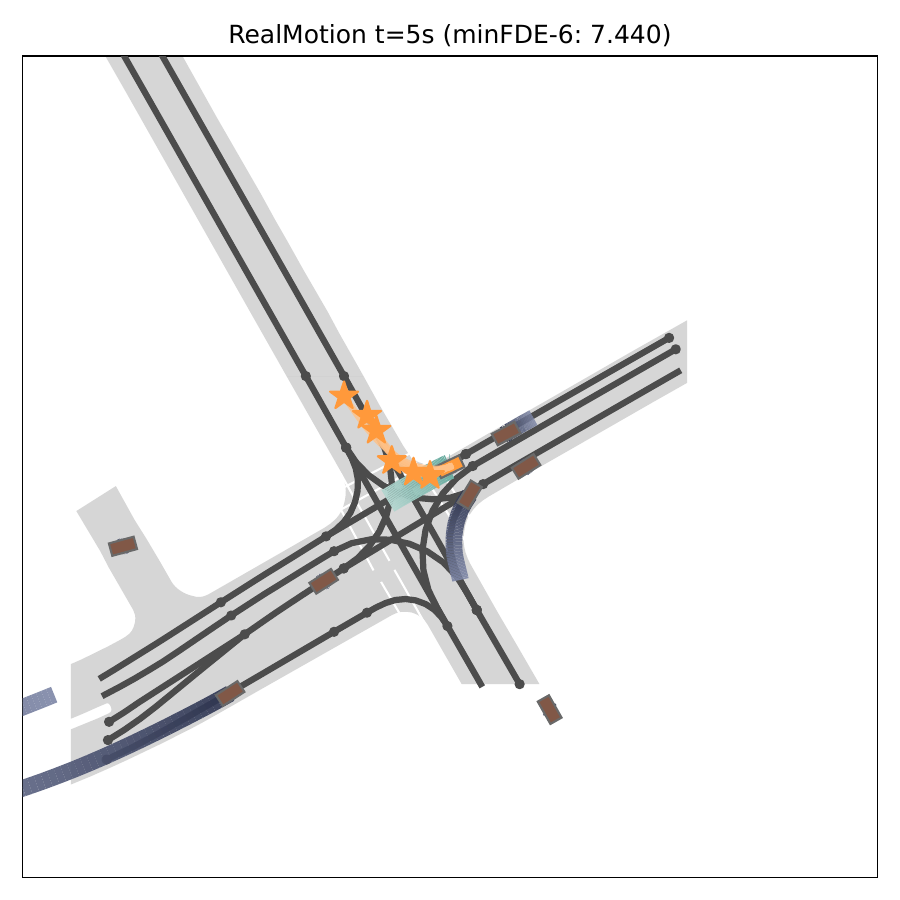}{2}{5}{4}{5}{0.025}{0.01}{1}
        \addresrow{3}{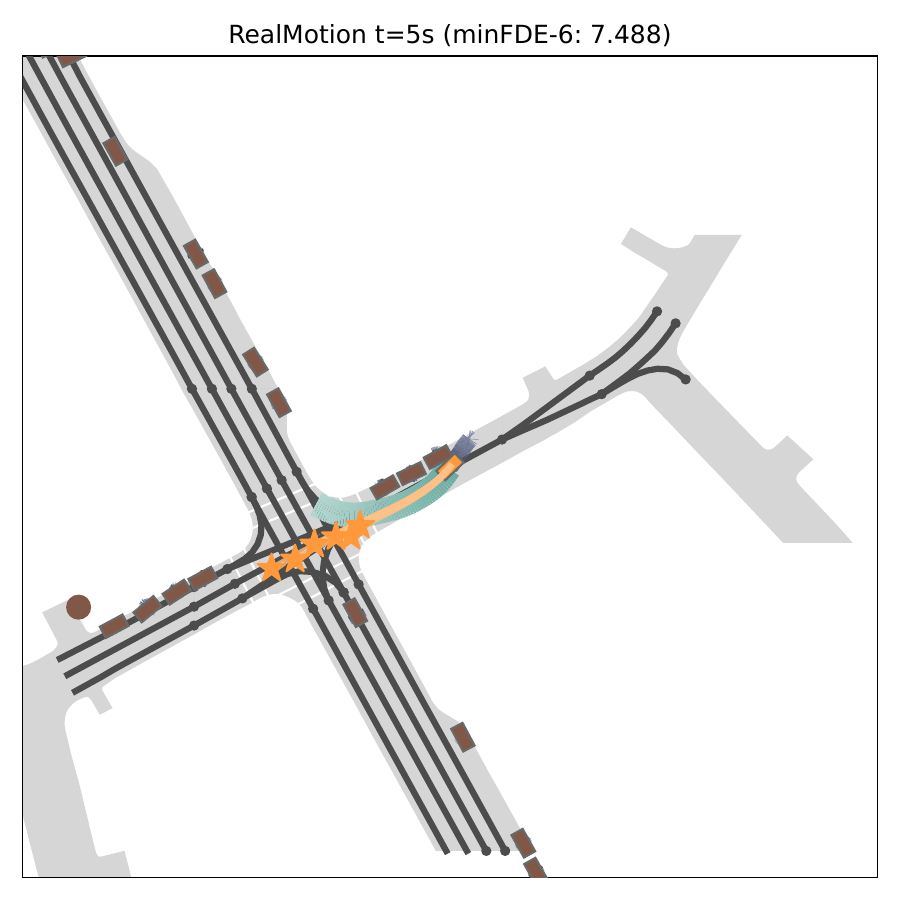}{1}{4}{3}{6}{0.025}{0.01}{2}
        \addresrow{4}{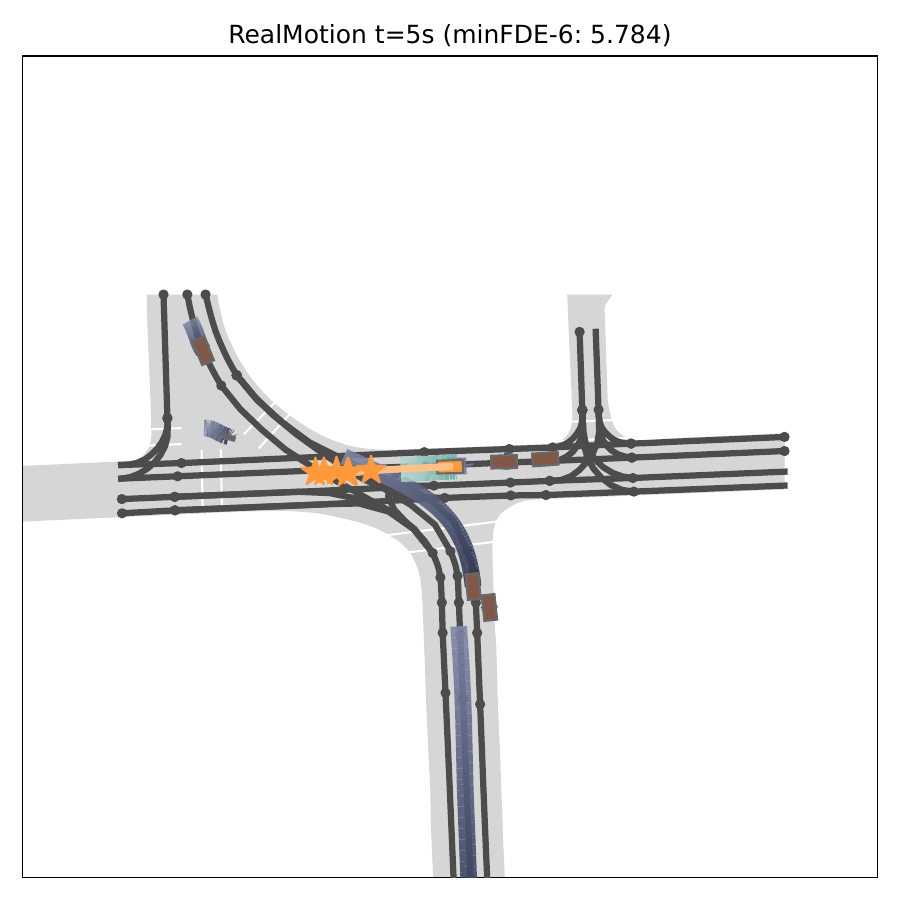}{1}{4}{3}{6}{0.025}{0.01}{3}
        \addresrow{5}{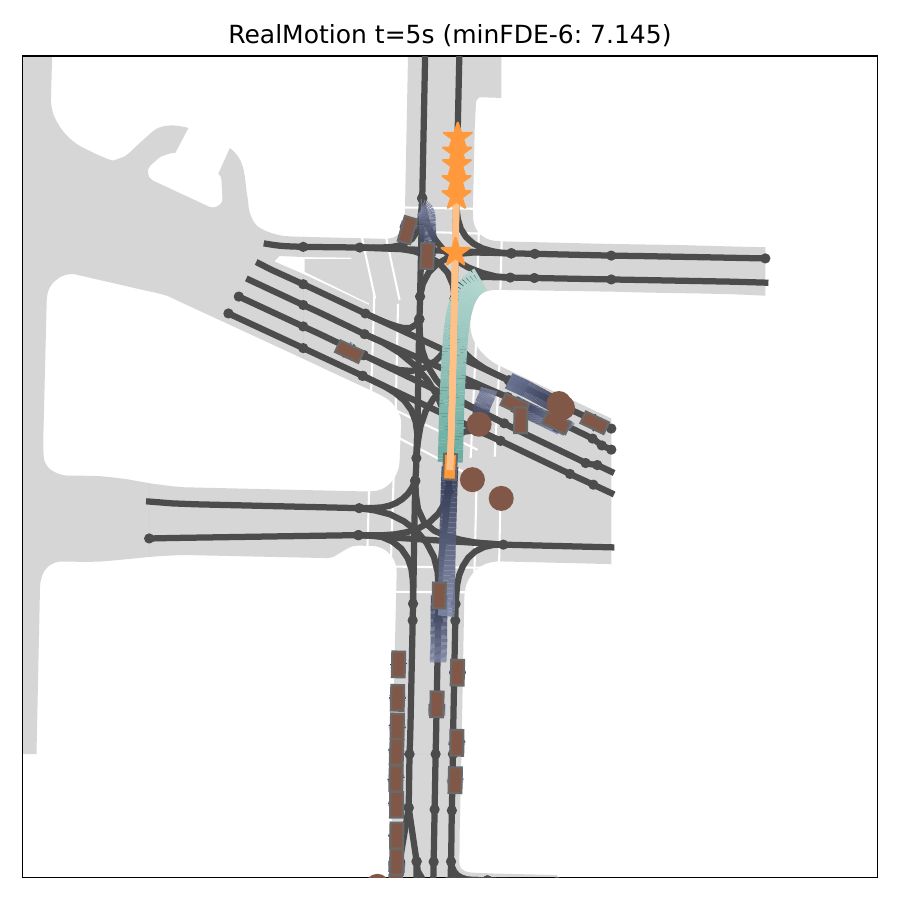}{2}{5}{3}{1}{0.025}{0.01}{4}
    \end{tikzpicture}
    }
    \caption{Qualitative results of our approach on scenarios from the Argoverse~2 validation set. We show the \textcolor[HTML]{ff9a3a}{\textbf{predictions}} of our streaming-based method at $t \in \{3, 4, 5\}$s.
    The visualizations also show \textcolor[rgb]{0.41, 0.67, 0.63}{\textbf{ground truth future}}, \textcolor[HTML]{384062}{\textbf{agent histories}}, and \textcolor[HTML]{815847}{\textbf{neighboring agents}}.
    The right column shows the final predictions at $t = 5$s for using RealMotion~\cite{song2024realmotion} in the streaming setting.}
    \label{fig:s_results2}
\end{figure*}

We present additional qualitative results on scenarios from the Argoverse~2 validation set in \Cref{fig:s_results1} and \Cref{fig:s_results2}.

\subsection{Failure Cases}
\begin{figure*}[tp]
    \centering
    \resizebox{\linewidth}{!}{
    \begin{tikzpicture}[node distance=0mm, inner sep=0mm]
        \addfailrow{0}{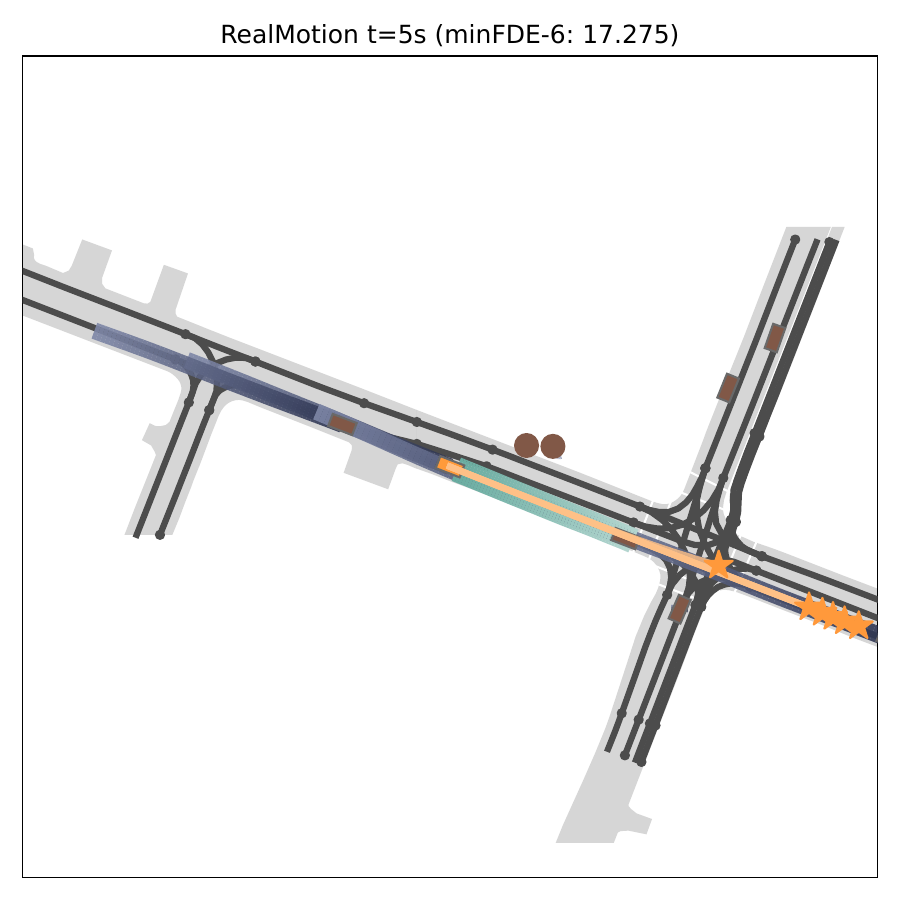}{5}{3}{0.5}{7}{0.025}{Unexpected stop}{}
        \addfailrow{1}{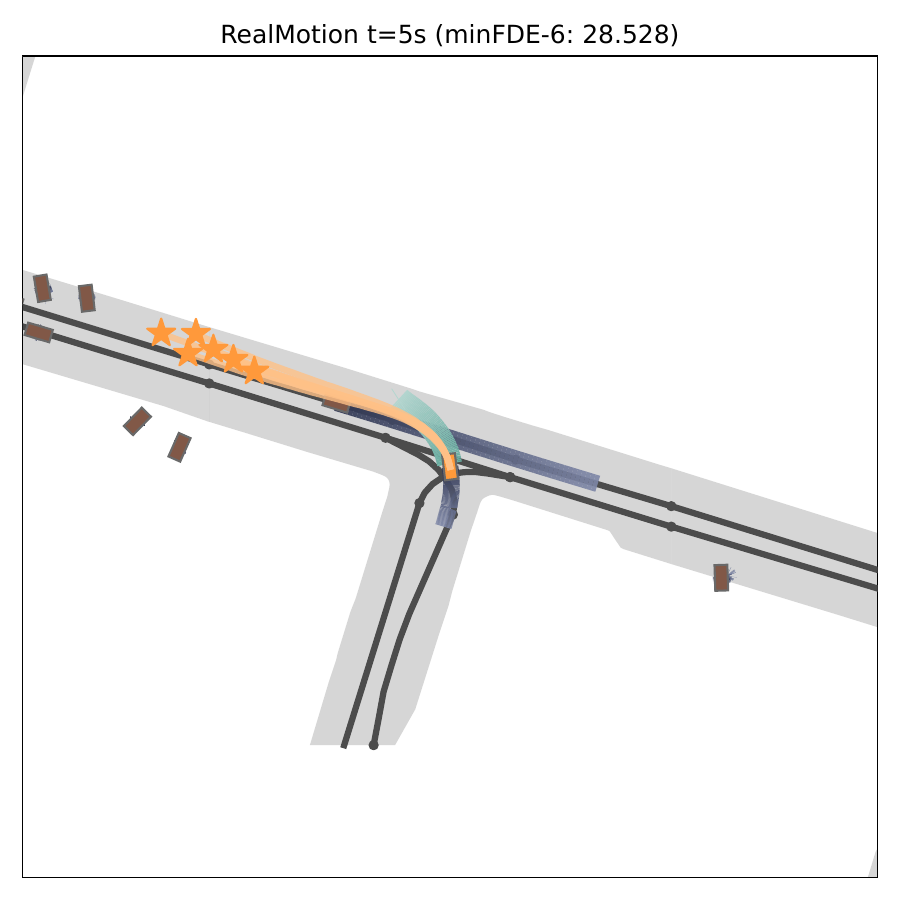}{0.5}{4}{5}{4}{0.025}{Unexpected stop}{0}
        \addfailrow{2}{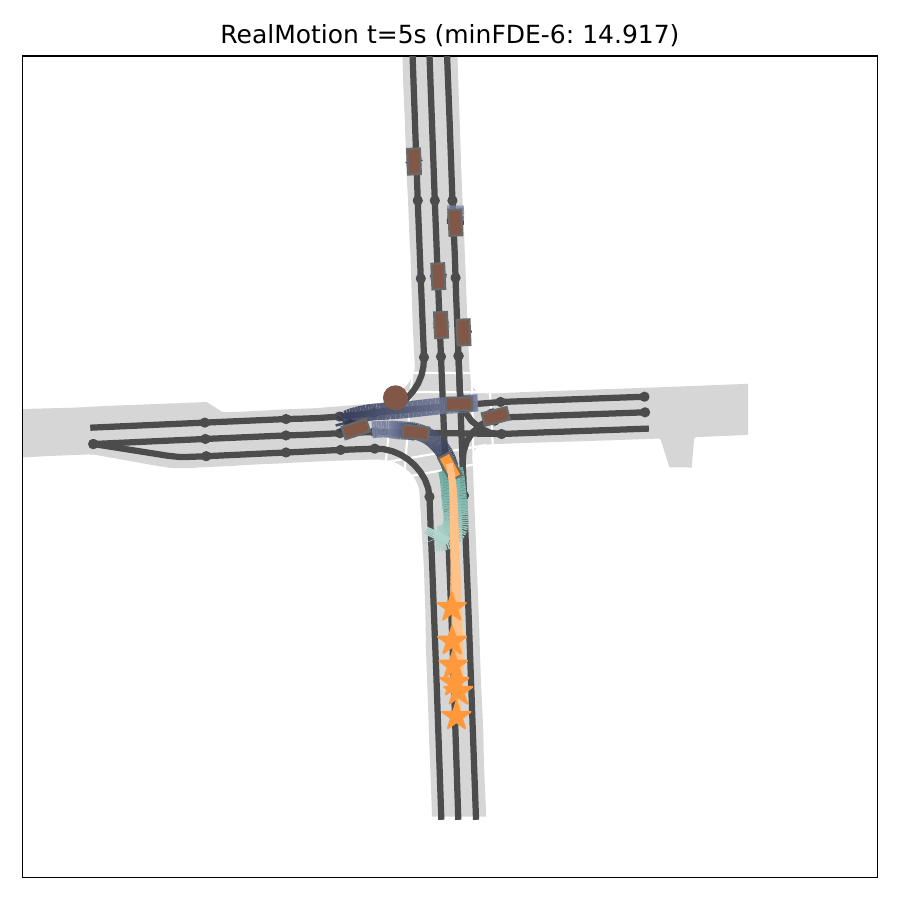}{3}{2}{4}{5.5}{0.025}{Driveway inadequately mapped}{1}
        \addfailrow{3}{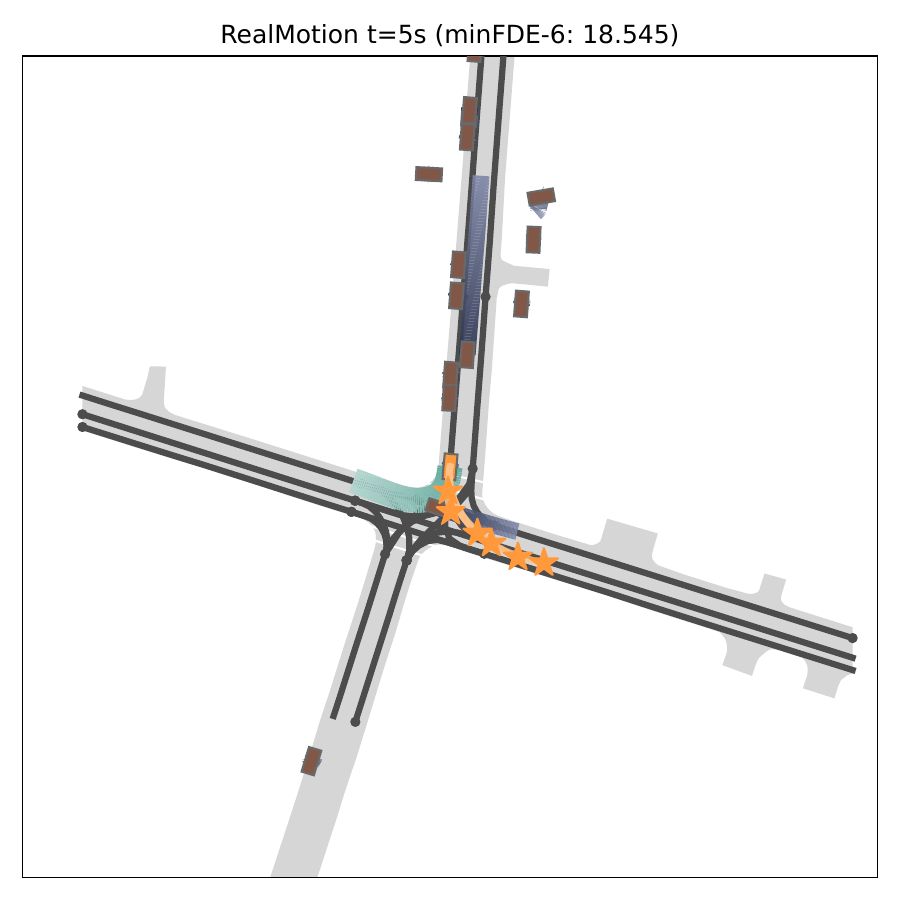}{3.5}{4}{3}{6}{0.025}{Waiting at intersection}{2}
        \addfailrow{4}{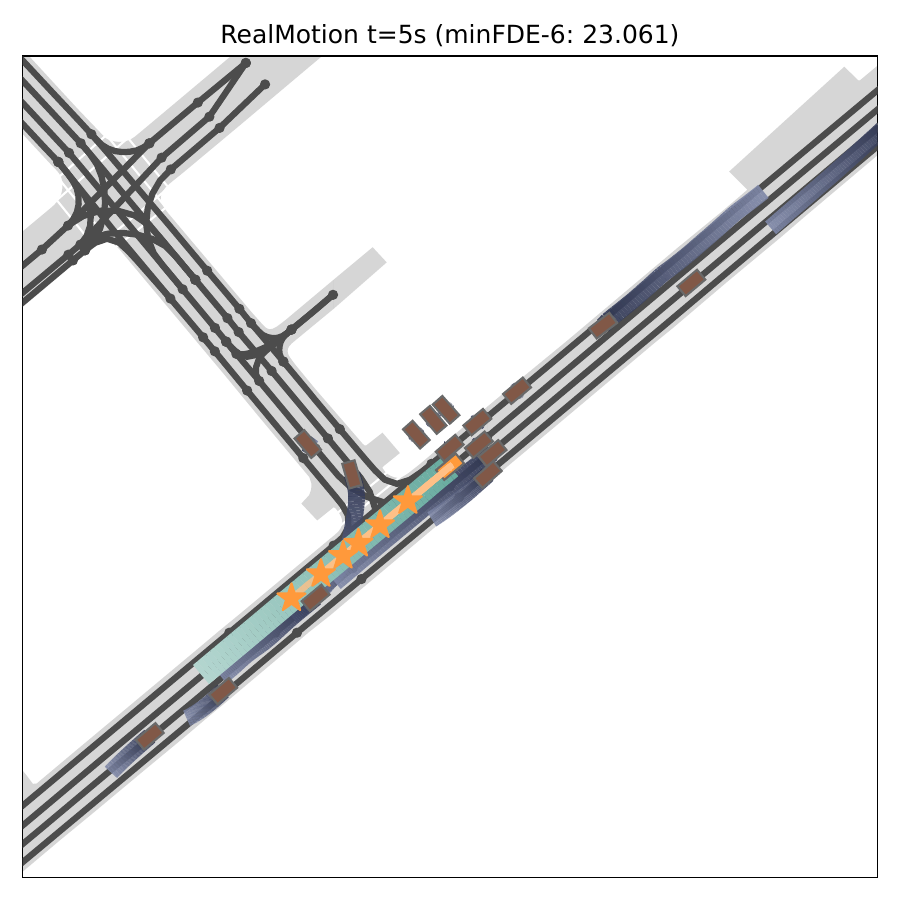}{0.5}{3.5}{6}{5.5}{0.025}{Waiting at intersection}{3}
        \addfailrow{5}{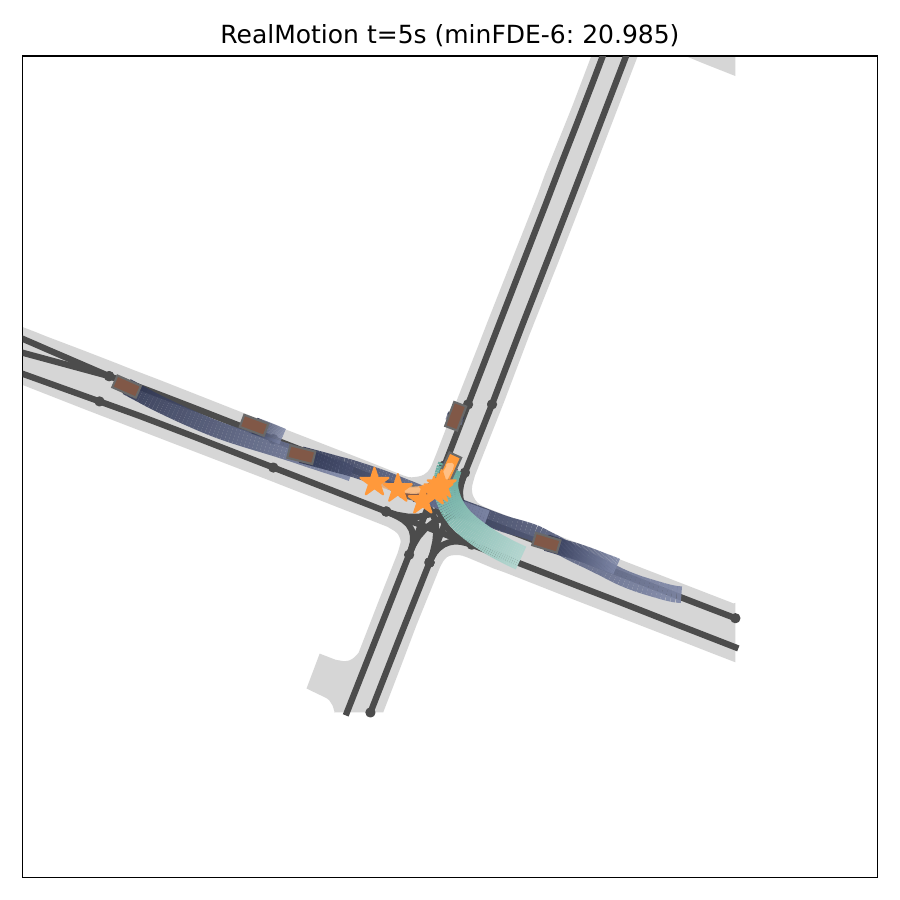}{3}{4.5}{5}{6}{0.025}{Waiting at intersection}{4}
    \end{tikzpicture}
    }
    \caption{
    Failure cases of our approach on scenarios from the Argoverse~2 validation set.
    We show the \textcolor[HTML]{ff9a3a}{\textbf{predictions}} of our streaming-based method at $t \in \{3, 4, 5\}$s.
    The visualizations also show \textcolor[rgb]{0.41, 0.67, 0.63}{\textbf{ground truth future}}, \textcolor[HTML]{384062}{\textbf{agent histories}}, and \textcolor[HTML]{815847}{\textbf{neighboring agents}}.
    The right column shows the final predictions at $t = 5$s for using RealMotion~\cite{song2024realmotion} in the streaming setting. \\
    In the first scenario, a vehicle stops before an intersection that is fairly far away for a currently unknown reason.
    In the second scenario, a vehicle stops for a reason that is not detected at the moment.
    In the third scenario, the vehicle begins turning into a driveway that is not modeled in the lane data.
    The fourth, fifth, and sixth scenarios depict vehicles waiting at an intersection where their future path is unclear.
    }
    \label{fig:s_failure1}
\end{figure*}

We present failure cases, where our approach fails to correctly predict future trajectories in \Cref{fig:s_failure1}.
Commonly, failures are introduced by agent movements which cannot be anticipated at the prediction time, often also due to inadequate map data, \ie missing modeling of driveways.

\section{Code Implementation}
\label{sec:app:code}
For better clarity and to facilitate reproducibility, we also provide our code implementation.
The accompanying \emph{ReadMe} file outlines how to setup a working environment and execute training, validation and visualization for the single-agent task.
The codebase also includes all implementations for the multi-agent setting (files marked with \texttt{ma} prefix or suffix).
We will release the source code and pretrained models upon paper acceptance.
Our code is based on the implementation of RealMotion\footref{fn:remo}.

\FloatBarrier

\end{document}